\documentclass[12pt]{article}

\usepackage{scicite}

\usepackage{times}

\usepackage{graphicx}%
\usepackage{multirow}%
\usepackage{amsmath,amssymb,amsfonts}%
\usepackage{amsthm}%
\usepackage{mathrsfs}%
\usepackage[title]{appendix}%
\usepackage{xcolor}%
\usepackage{textcomp}%
\usepackage{float}
\usepackage{paralist}
\usepackage{color}
\usepackage{xcolor}
\usepackage{xxcolor}
\usepackage{pgf}
\usepackage{pgfplots}
\usepackage{tikz}
\usepackage{setspace}
\usepackage[skins]{tcolorbox}
\usepackage{xspace}
\usepackage{todonotes}
\usepackage{url}
\usepackage{adjustbox}
\usepackage{algorithm}
\usepackage{algpseudocode}
\usepackage{afterpage}

\algnewcommand{\IfThenElse}[3]{% \IfThenElse{<if>}{<then>}{<else>}
  \State \algorithmicif\ #1\ \algorithmicthen\ #2\ \algorithmicelse\ #3}
\algnewcommand\algorithmicforeach{\textbf{for each}}
\algdef{S}[FOR]{ForEach}[1]{\algorithmicforeach\ #1\ \algorithmicdo}

\usetikzlibrary{positioning}
\usetikzlibrary{calc}
\usepgfplotslibrary{groupplots,fillbetween,}
\usetikzlibrary{arrows,shadows,petri,automata,shapes.geometric,shapes,patterns}

\usepackage{caption}
\newenvironment{sciabstract}{%
\begin{quote} \bf}
{\end{quote}}

\title{Hierarchies define the scalability of robot swarms}

\author
{Vivek Shankar Varadharajan$^{1\ast}$, 	
Karthik Soma$^{1}$, Sepand Dyanatkar$^{1,2}$,\\ 
Pierre-Yves Lajoie$^{1}$, Giovanni Beltrame$^{1}$\\
\\
\normalsize{$^{1}$Department of Computer and Software Engineering, Polytechnique Montreal,}\\
\normalsize{Canada}\\
\normalsize{$^{2}$Department of Computer Science, University of Cambridge,}\\
\normalsize{United Kingdom}\\
\normalsize{$^\ast$ E-mail: giovanni.beltrame@polymtl.ca}
}

\date{}

\begin{document} 


\baselineskip24pt


\maketitle 

\noindent\textbf{Short title:} \textit{Hierarchical robot swarms}


\begin{sciabstract}
    The emerging behaviors of swarms have fascinated scientists and
  gathered significant interest in the field of robotics. Traditionally, swarms
  are viewed as egalitarian, with robots sharing identical roles and
  capabilities. However, recent findings highlight the importance of
  \emph{hierarchy} for deploying robot swarms more effectively in diverse
  scenarios.

  Despite nature's preference for hierarchies, the robotics field has clung to
  the egalitarian model, partly due to a lack of empirical evidence for the
  conditions favoring hierarchies. Our research demonstrates that while
  egalitarian swarms excel in environments proportionate to their collective
  sensing abilities, they struggle in larger or more complex settings.
  Hierarchical swarms, conversely, extend their sensing reach efficiently,
  proving successful in larger, more unstructured environments with fewer
  resources. We validated these concepts through simulations and physical robot
  experiments, using a complex radiation cleanup task. This study paves the way
  for developing adaptable, hierarchical swarm systems applicable in areas like
  planetary exploration and autonomous vehicles. Moreover, these insights could
  deepen our understanding of hierarchical structures in biological organisms.
\end{sciabstract}

\noindent\textbf{Summary:} \textit{Hierarchical structures allow robot swarms to scale in size and task
complexity in a cost-effective manner}

\section*{Introduction}\label{main}

Robot swarms are large-scale multi-robot systems controlled by simple rules
which combine into complex, self-organised behaviors\cite{Dorigo2021}. Despite
their potential, robot swarm face important challenges
such as the availability of reliable state estimation, the complexity of
behavior design, and the limited sensing capabilities of simple robots.

Introducing a hierarchical structure is a promising way to unlock robot swarms'
potential\cite{dorigo2020reflections}. In such hierarchies, certain swarm
members have advanced capabilities or knowledge, and can guide collective
decision-making. This structure aids in information distribution and
aggregation, addressing issues like congestion in dense
swarms\cite{soma2023congestion}, and streamlines the allocation of
responsibilities, allowing specialized robots to assume leadership roles, and
ultimately lower the overall swarm cost.

Traditionally, swarm robotics research has focused on how simple,
nature-inspired control laws lead to coordinated
behavior\cite{Rubenstein2014,Slavkov2018,boudet2021collections,Vasarhelyi2018,li2019particle,Berlinger2021,McGuire2019}.
While some studies have explored leader-follower dynamics\cite{Tron2016,
  consolini2008leader, panagou2014cooperative, ji2006leader}, they often result
in followers merely replicating leaders' actions, and not in multi-level
decision-making. Other research has looked at the emergence of
specialization\cite{halasz2013emergence,li2002emergent}, but the field
generally prioritizes egalitarian swarms over the hierarchical structures
frequently seen in nature\cite{Nagy2010,flack2018local,jiang2017identifying,
  seeley2014honeybee,giuggioli2015delayed,peterson2002leadership}, like in bird
migration\cite{flack2018local} or primate social
systems\cite{Baboons,wittig2003food}.

% Figure 1
\begin{figure}[H]
\includegraphics[width=\textwidth]{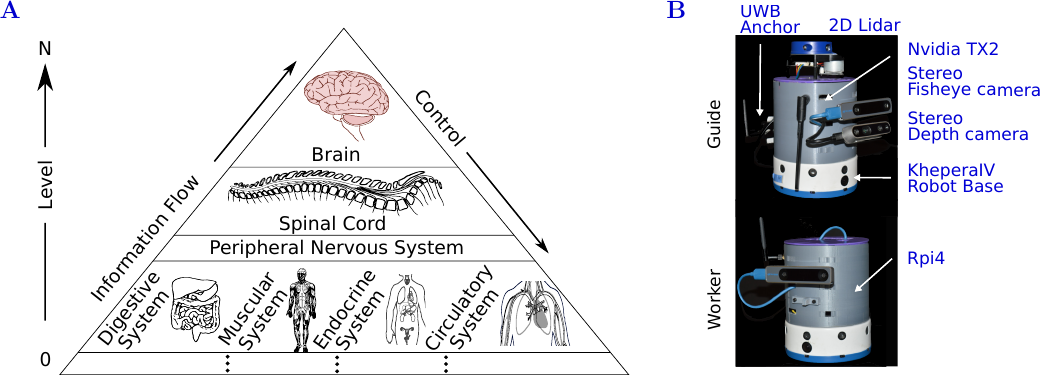}
\caption{Hierarchies in robot swarms. (A) The hierarchy represented by the
  organs in human biology, the brain as a whole performs a complex task
  (maintaining the well-being of the individual), receives sensory information,
  and controls the individual organ system through its nervous
  system and down to a microscopic scale each organ is made of similar cells.
  (B) The two types of robots used in the hierarchy: Guide robots are equipped with long range sensing and sophisticated on-board computer, and the worker robots are simple robots with low-cost computation and proximity sensors. 
  %(C) Illustration of various behaviors
  %performed by the robots during the prototype mission: guide robots perform
  %systematic exploration to identify the target location (1a), and workers use a
  %bug-inspired algorithm\cite{McGuire2019} (1b-1d) to identify targets. Robot guides
  %communicate with other guides in range once they have identified a target, and
  %navigate home (2a) to support workers to the target using chain formation (4).
  %Worker robots that did not find the target return home using a home beacon
  %(2b). Workers act as beacons for other workers once they have identified a
  %target, attracting other workers (3). The overall mission of the robots is to
  %identify the target, and navigate all the robots to its location.
%   The map generated by the most updated guide robot during one of the experimental run employing Hierarchical strategy in urban (C), maze (D), and forest (E). (F-H) show the visitation frequency heat map of the worker robot when using egalitarian strategy during three repetitions in urban (F), and maze (G), the forest (H) with single repetition as new forests are generated for each run.   
}
\label{fig:illus}
\end{figure}

Egalitarian swarms require robots counts that are proportional to the complexity
of environment or task, but face diminishing returns as the number of robots
increases\cite{hamann2018swarm,soma2023congestion}. We posit that \textbf{a
  hierarchical approach, where select swarm members perceive the global state
  and objectives, can scale robot swarms cost-effectively}. This
hierarchy, whether emerging\cite{zhu2024selforganizing} (as in brain function,
see Figure~\ref{fig:illus} A) or designed (like heterogeneous robot
swarms\cite{Mathews2017}), aggregates local information for tiered
decision-making, avoiding expensive or impractical overdesign.

Notably, hierarchies implicitly exist in deployed systems like those of the
DARPA Subterranean
Challenge\cite{CERBERUS,agha2021nebula,hudson2021_CSIRO61,ohradzansky2021_marble},
where decentralized autonomous systems are overseen by human operators. We
mirror this supervised autonomy but distribute control across different swarm
levels instead of having a single centralized entity. With top-level swarms
handling group tasks and lower levels focusing on individual robot roles.

%\subsubsection*{Experimental Setup}

To validate our theories, we developed a radiation cleanup mission, consisting
of locating radiation targets and deploying a minimum number of robots for
cleanup. We evaluated 3 control strategies using the two robot types shown in
Figure~\ref{fig:illus} B:
\begin{compactenum}
\item Egalitarian: radiation-tolerant robot ``workers'' employ reactive
  behaviors to explore and act as beacons to attract nearby robots, forming
  teams that reach the target locations. Workers can use a ``bug'' exploration
  algorithm to find the targets\cite{McGuire2019} or follow a beacon signal to
  return to their initial location.
\item Hierarchical: workers rely on costly, advanced but
  radiation-sensitive\cite{ezell2021radiation} ``guides'' to reach their target
  locations. Guides explore and identify target with advanced sensors,
  subsequently leading the workers to the targets.
\item Heterogeneous: workers and guides explore simultaneously. Target detection
  information is shared among all robots.
\end{compactenum}
All robots can detect proximity to radiation and communicate with other robots
within a short communication range. A mission is successful if all targets are
identified and a minimum of 10 worker robots are assembled at each location.
Missions are to be completed in the shortest possible time.
Figure~\ref{fig:state_machine} shows the states used by
workers and guides to complete the mission.

\afterpage{%
\begin{figure}[H]
\centering
\includegraphics[width=0.84\textwidth]{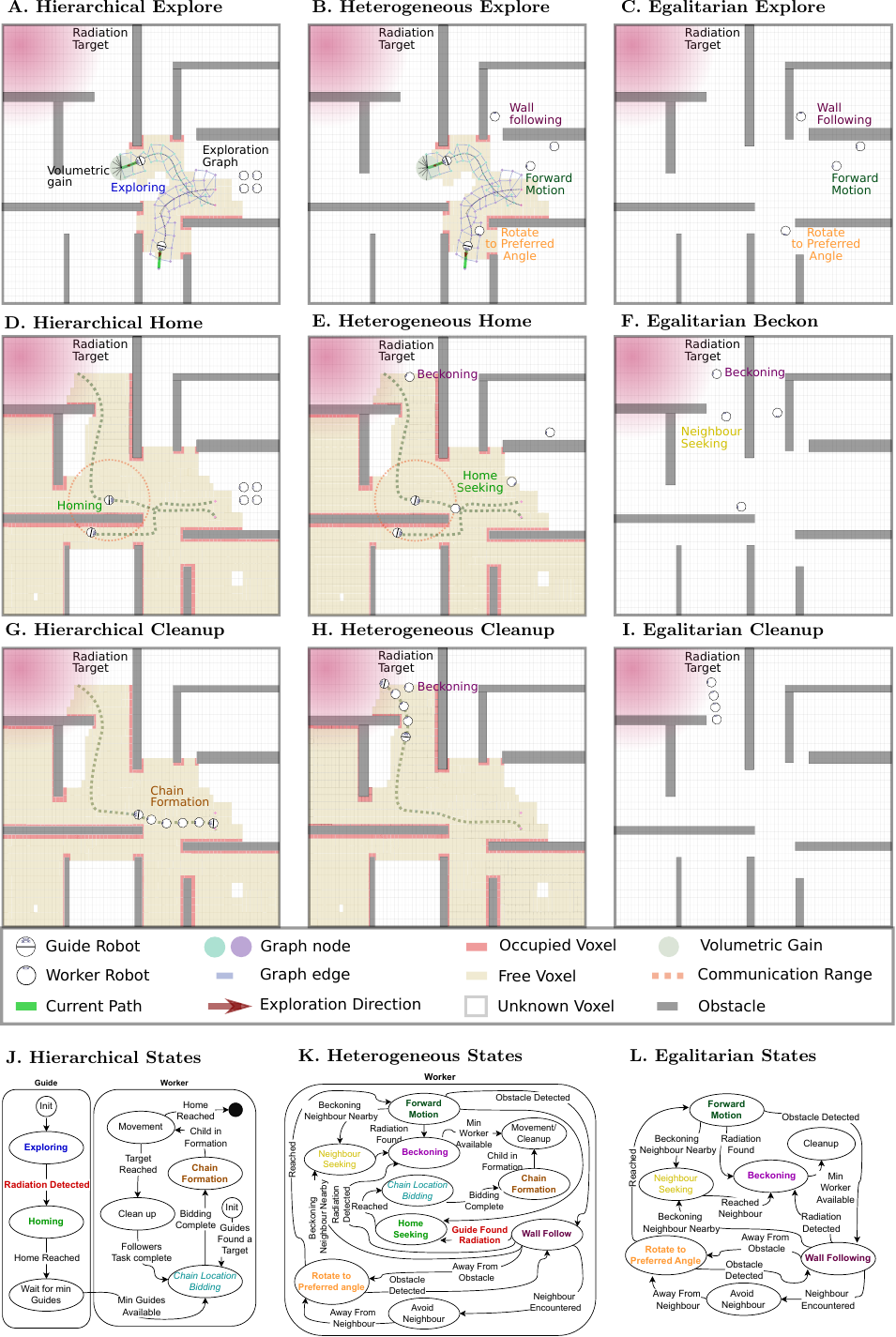}
\caption{The set of behaviors performed by the robots during the three control
  approaches: In Hierarchical, the guides explore to identify the targets (A),
  on identification of the target they return home (D), and form a chain to
  transport the workers to the target (G). During Heterogeneous control, the
  exploration is performed by both guides and workers using their sensing
  abilities (B), on identification of target guides return home with broadcasts
  of target identification, workers receiving the broadcasts return home (E),
  and the workers are transported to the target in chain formation (H). In
  Egalitarian, all the robots explore using bug-inspired
  algorithm\cite{McGuire2019} (C), robot create beacons around the target to
  attract other bugs (F), and the radiation cleanup task is performed after the
  required number of robots reach the target (I). The overall mission of the
  robots is to identify the target, and navigate all the robots to its location.
  The state machine in (J-L) shows the sequential states performed by the guide
  and worker robots, respectively during Hierarchical, Heterogeneous and
  Egalitarian. }
\thispagestyle{empty}
\label{fig:state_machine}
\clearpage
\end{figure}}

We believe this task is particularly suitable for the study of hierarchies:
\begin{inparaenum}
\item it simulates a communication-restricted environment where robots can only
  communicate when in close proximity, mirroring many biological systems,
  physical processes, and real-world scenarios;
\item it demands spatial awareness for target localization;
\item it involves allocating a specific number of resources (e.g., workers) to
  particular cleanup tasks;
\item it is representative of many typical swarm applications, such as
  search and rescue, prospecting, foraging, and others;
\item the environment's susceptibility to congestion~\cite{soma2023congestion}
  is minimal, as densely populating it with robots positions them closer to
  their targets, enabling a direct comparison of hierarchical cost scalability
  under challenging conditions—specifically, when egalitarian swarms can expand
  without restriction.
\end{inparaenum}

We conducted physics-based simulations across three types of environments
(Urban, Maze, and Forest, showcased in supplementary Figures~\ref{fig:map} D-F) of varying
size with radiation targets located on the periphery, as well as real robot
experiments. Further details are available in the Methods and Supplementary
Material.

\section*{Results}

\subsection*{Limits of Egalitarian Swarms}

\begin{figure}[tpbh]
\includegraphics[width=\textwidth]{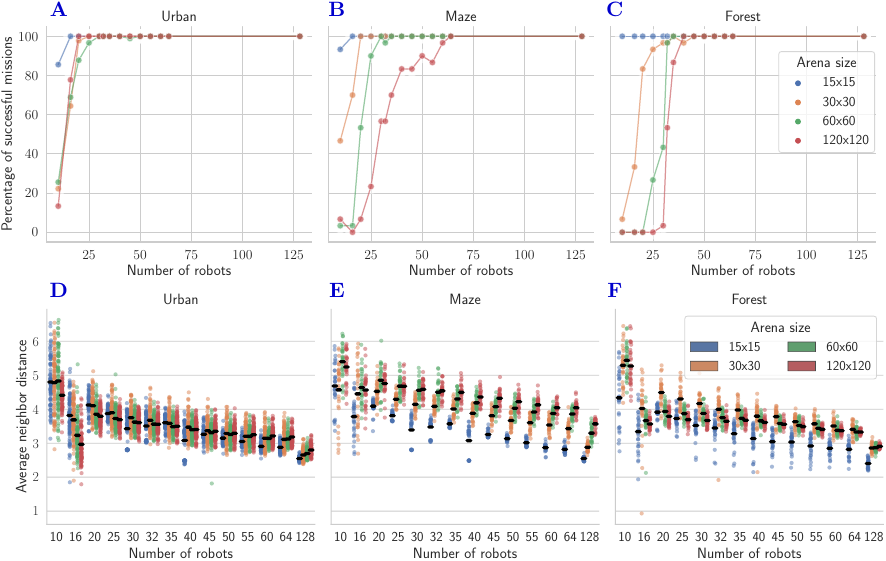}
\caption{The percentage of successful missions when scaling the number of worker robots and map size in urban (A), maze (B), and forest (C). Average neighbour distance for all arena configurations are shown in (D-F). As the number of robots increase the success rate improves which is synchronous to the drop in average neighbour distance, indicating that sufficient collective sensing area is required to succeed in the mission. }
\label{fig:sim_results1}
\end{figure}

\begin{figure}[tpbh]
\includegraphics[width=\textwidth]{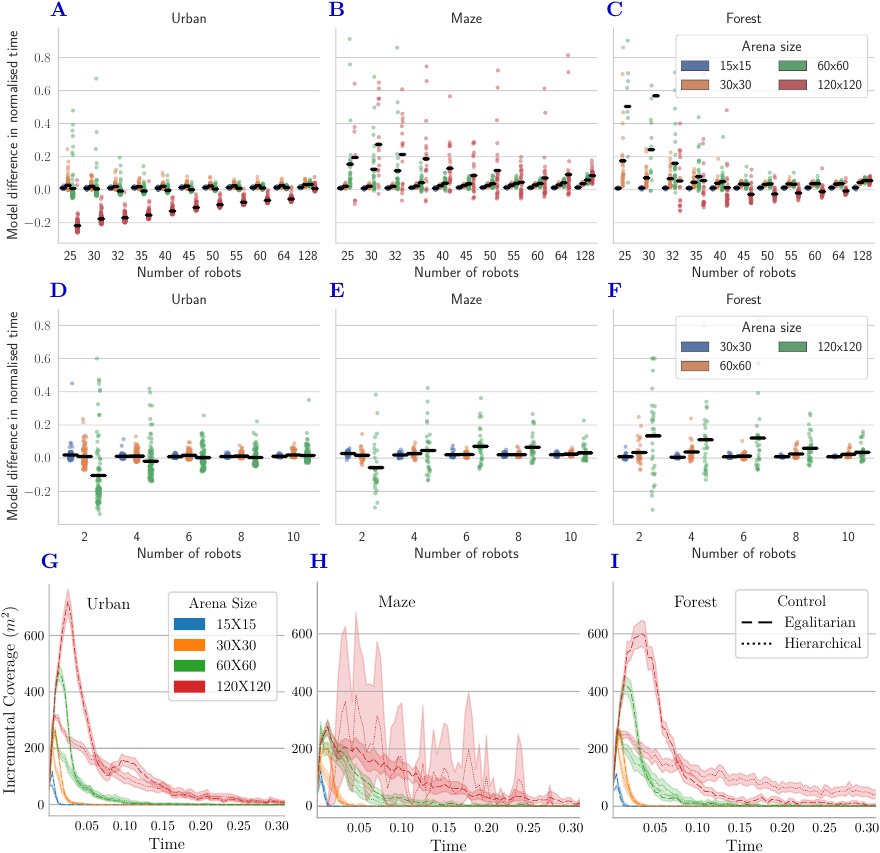}
\caption{The difference in Normalised time between the model and the empirical outcome of the egalitarian strategy in urban (A), maze (B), and forest (C) with an increasing number of robots. The model estimates in all three environments are closer to the empirical outcome with either a larger number of robots or smaller environments. The empirical outcome gets closer to the model when collective sensing of the swarm is sufficient for the arena's size, indicated by the percentage of successful mission completion. The model overestimates the completion time in larger urban environments with fewer robots. In larger maze environments with fewer robots, the model underestimates the mission completion time. Forest environments with fewer robots tend to consume time over the model estimates and drop below the model estimated time as the number of robots increases. The model differences occur due to obstacle structures that either lead a robot to the target, as with urban environments or delay the robot in reaching a target, as in the maze. The difference in Normalized time between the model and the empirical outcome when using hierarchical control in urban (D), maze (E), and forest (F) with increasing guides. As the number of guide robots increases, the difference in time between the model and the empirical outcome reduces in all arena configurations. The difference between the model and the empirical outcome gets more pronounced with larger environments. (G-I) Mean incremental coverage and its 95\% confidence interval with egalitarian and hierarchical control in urban (G), maze(H), and forest (I) with all area sizes. The incremental coverage of
  egalitarian starts off large due to a large number of robots exploring in
  parallel but fades off to be lower than hierarchical quickly when the mission
  leads to a failure, particularly in large forest and maze environments.} 
\label{fig:model_diff}
\end{figure}

Figures~\ref{fig:sim_results1} A-C show the success rates of Egalitarian
swarms in various environments, sizes, and robot counts. The trend is clear:
success rates reach 100\% as the number of robots increases, aligning with the
Poisson germ-grain model's predictions\cite{chiu2013stochastic}. This model
relates swarm success to factors like sensing radius ($s_{r}$), velocity
($v_{r}$), and density ($\rho$). Model predictions versus actual outcomes are
shown in Figures~\ref{fig:model_diff} A-F.

% Figure 3

% Figure 4

The combined effect of $\rho$, $s_{r}$, and $v_{r}$ must exceed a threshold
related to the complexity of the environment to achieve target detection within
a finite time. We define this concept the \emph{swarm sensing area}, reflecting
the swarm's collective detection capability. Low $s_{r}$ means robots must move
more to discover targets, while low $\rho$ hinders information sharing within
the swarm. However, higher $v_r$ can compensate by speeding up exploration and
allowing more frequent communication.

Robot velocity and sensing radius are generally bound by hardware limitations:
improving these aspects can be difficult and would raise cost per robot. Adding
robots boosts $\rho$, effectively extending the swarm's sensing area (see
Figures~\ref{fig:sim_results2} I-J). Enhancing $s_{r}$ similarly boosts
performance, supporting the concept of the swarm sensing area (see supplementary
Figure~\ref{fig:bug_doubling_sensing_range}).

Figures~\ref{fig:sim_results1} D-F display the average distance between
neighbouring robots as the number of workers increases. Combined with the success
rate in Figures~\ref{fig:sim_results1} A-C, the trend suggests that consistent
mission success correlates with an average distance below 2x the sensing range
of the robots.

\subsection*{Advantages of Hierarchical Swarms}

\begin{figure}[H]
\centering
\includegraphics[width=\textwidth]{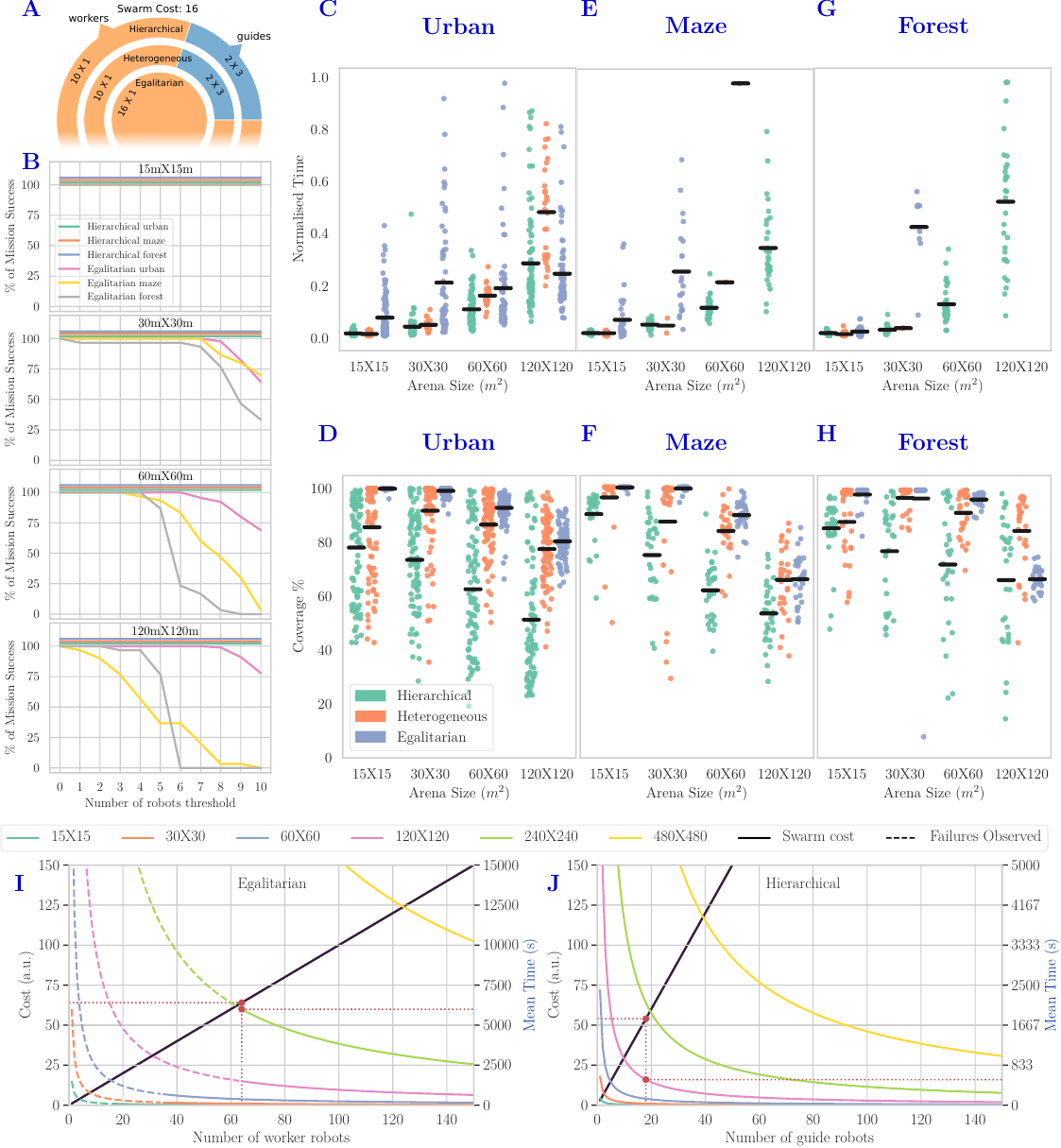}
\caption{Comparison between Hierarchical, Heterogeneous, and Egalitarian
  approaches. (A) shows the cost of individual robots with the three approaches.
  (B) The percentage of successful missions versus the number of robots required
  to find the target to succeed with a hierarchical and egalitarian strategy in
  various experimental configurations. The normalized time taken to complete the
  mission with the three approaches for urban, maze, and forest are shown in
  (C), (E), and (G), respectively (missing data points correspond to mission
  failure). (D,F and H) show the areas explored for Urban, Maze, and Forest
  environments, respectively. The black bars in (C,E and G) and (D,F, and H)
  indicate the mean of the data cloud. (I-J) shows the swarm cost to model based
  time performance of egalitarian and hierarchical strategy over different arena
  sizes, and the right y-axis shows the meantime and left swarm cost. For
  instance, take the red points; a swarm cost of 64 with an egalitarian strategy
  might take 6000s, and a cost of 64 with 18 guides and 10 workers in the
  hierarchy might take 533.33s to complete the mission in a 120x120m
  environment. The model based projections could be used in system design as the
  empirical outcome closely follow the model.}
\label{fig:sim_results2}
\end{figure}

Figure~\ref{fig:sim_results2} B indicates that the Hierarchical approach
consistently achieves 100\% success for the fixed-cost swarm in
Figure~\ref{fig:sim_results2} A. 

%The Egalitarian and
%Heterogeneous approaches mainly succeed in smaller areas or when reducing the
%number of required robots at targets. Guides in a hierarchy explore
%systematically, mapping the environment and synchronizing knowledge when in
%range, as detailed in the Supplementary Material. Figures~\ref{fig:illus} C-E
%and F-H show the guides' synchronized maps and worker robots' visitation
%heatmaps, respectively.

% Figure 5

%In the Hierarchical setting, once a guide locates a target, it leads workers
%there directly. In contrast, Egalitarian control lacks the ability to share
%target coordinates, causing extensive area coverage without necessarily
%achieving mission success (Figures~\ref{fig:sim_results2} D, F, H).

Figures~\ref{fig:sim_results2} C, E, G show mission times for fixed-cost swarms,
with completion time increasing exponentially in larger arenas due to extended
exploration. This growth pattern aligns with the Poisson process model
(Figures~\ref{fig:sim_results2} I, J) and could serve as a template for system
design to compute the optimal number of robots for a given mission.

The Heterogeneous approach uses all robots to explore in parallel, but despite
its rapid discovery of the targets (see supplementary Figure~\ref{fig:first_target_detection}), its
effectiveness is limited as many workers get stuck while exploring and do not
reach the targets, even with the assistance of the guides
(Figures~\ref{fig:sim_results2} C,E,G).

Area coverage analysis (Figures~\ref{fig:sim_results2} D,F,H and
Figures~\ref{fig:model_diff} G-I) reveals a decrease in explored area with
larger environments for the Hierarchical approach. However, Egalitarian and
Heterogeneous approaches do not always translate larger explored areas into
success due to challenges in delivering robots to targets. Hierarchical methods
steadily increase coverage until all targets are found, while Egalitarian and
Heterogeneous have an initial burst of exploration followed by stagnation:
workers do not maintain knowledge of previously explored regions or target
locations and often become trapped in cycles.

\begin{figure}[H]
\includegraphics[width=\textwidth]{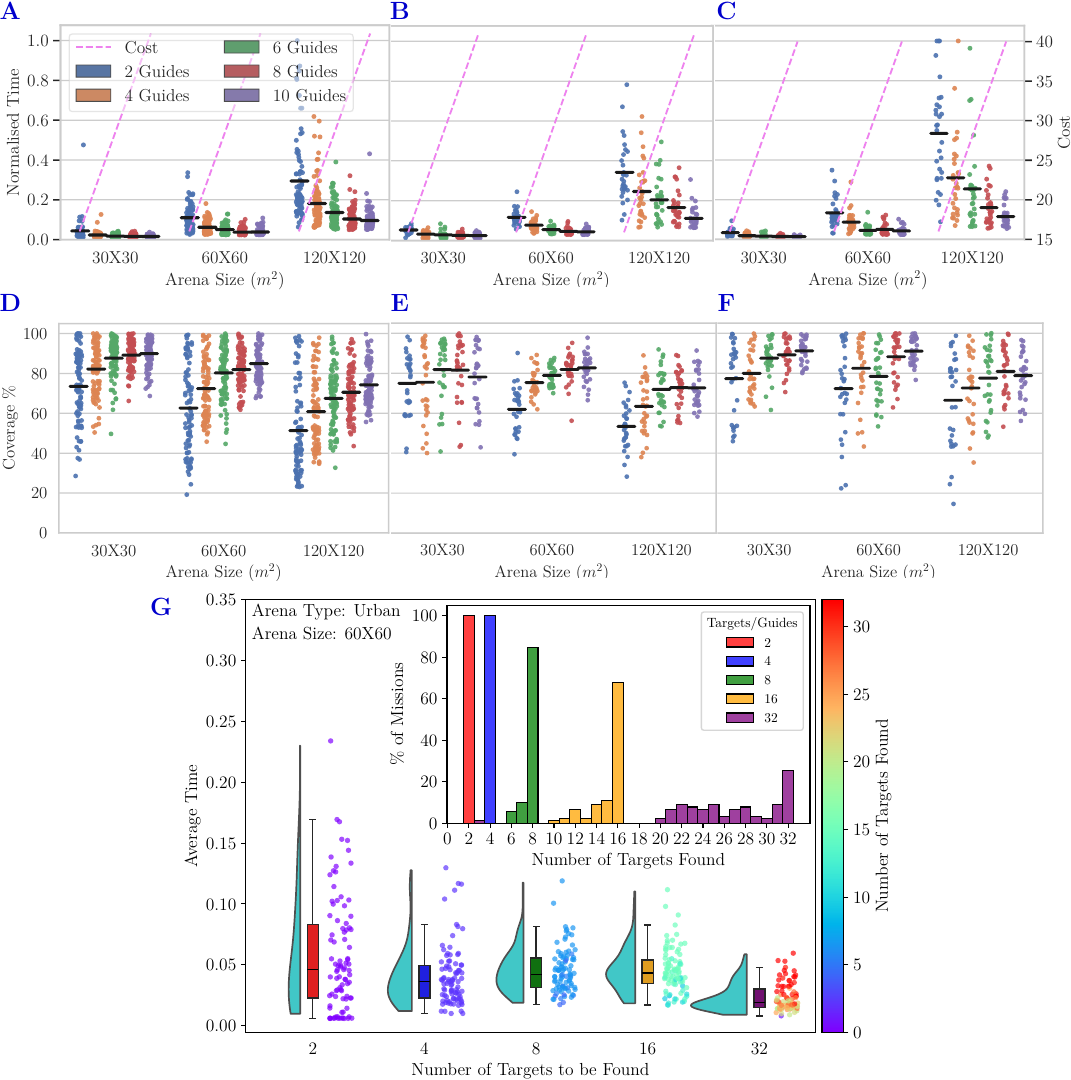}
\caption{Normalised time taken in simulation (A-C) and the amount of explored area (D-F)
  with a varying number of guides in a hierarchical approach for Urban, Maze,
  and Forest environments with a single target. Average time to find the targets
  when changing the number of targets (G) in the 60 X 60 urban environment.
  }
\label{fig:sim_scalability}
\end{figure}

Figures~\ref{fig:sim_scalability} A-C and D-F display mission time and explored
areas when increasing the guides count, following the Poisson model
(Figure~\ref{fig:sim_results2} J). Figure~\ref{fig:sim_scalability} G shows
consistent average assembly times at targets irrespective of the number of
targets and guides (with some variability due to congestion) suggesting
successful parallelization of the worker delivery to the different target sites.
We report statistically significant differences between Hierarchical and
Egalitarian approaches (see supplementary Tables 6-8).

% Figure 6

\begin{figure}[H]
\includegraphics[width=\textwidth]{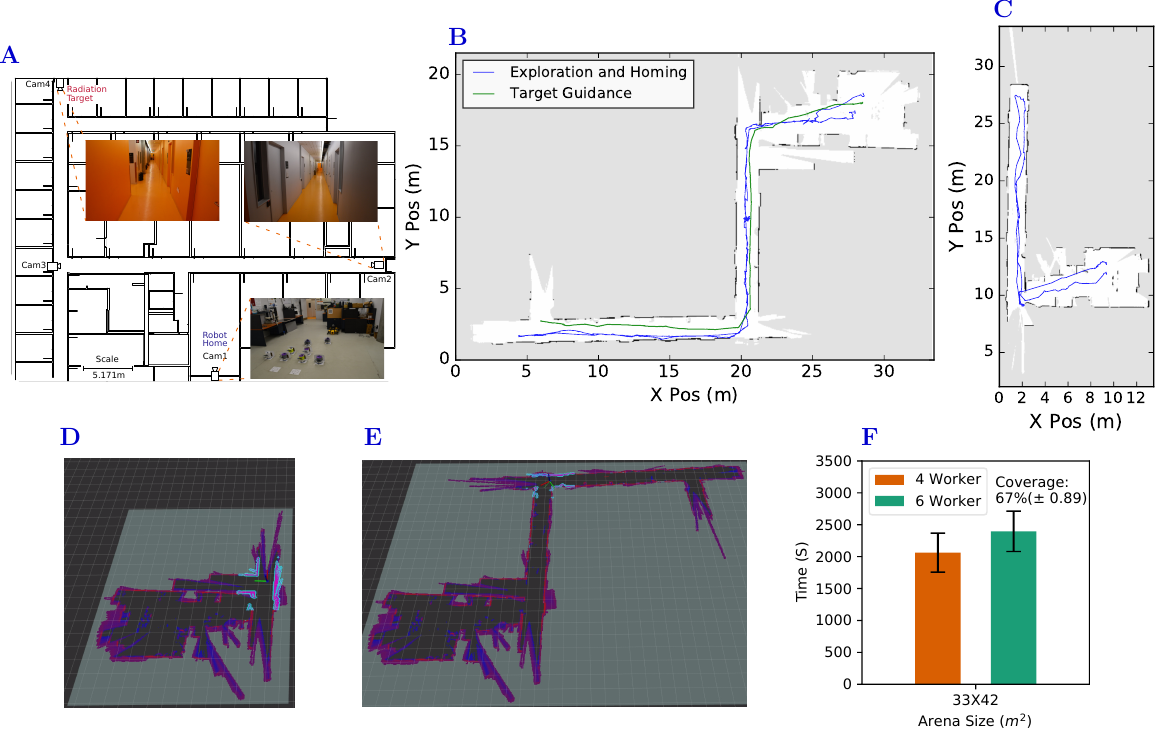}
\caption{Experimental setup of the physical robot experiments:(A) the floor plan
  of the section of the experimental arena. Experimental results of the physical
  robot experiments: (B) the map generated by one of the two guide robots that
  found the target along with the trajectories taken during exploration, return
  to base and worker delivery, (C) the map generated by the other guide robot,
  (D) the base station visualization with the exploration trajectory, and (E)
  base station view during homing. (F) the time taken for delivering 4 and 6
  worker robots.}
\label{fig:realrobot}
\end{figure}

Real-world tests (see Figures~\ref{fig:realrobot} A-F) validate the simulation
findings with matching coverage and mission completion times, accounting for the
simulation ``reality gap''.

% Figure 7 

\section*{Discussion and Outlook}  

% The primary focus of our research is to evaluate the performance of a
% hierarchically organized robot swarm compared to an egalitarian swarm.
% Hierarchies introduce selected swarm members with enhanced capabilities,
% offering a broader perspective to guide the swarm towards its mission
% objectives.

Results suggest that hierarchies are cost-effective deployment strategies. For
instance, 100\% success rate in the largest environment requires 64 Egalitarian
robots versus only 12 in a Hierarchical setup (2 guides, 10 workers), a 400\%
cost reduction. Our experiments also highlight:
\begin{compactenum}
\item Reactive behaviors require a swarm sensing area proportional to the
  environment size, as shown in Figures~\ref{fig:sim_results1} A-C. This
  requirement stems from the swarm's lack of memory and the need for high
  density for information propagation. Endowing a small subset of
  robots with spatial awareness ensures a stable success rate
  regardless of swarm size, as evidenced in Figure~\ref{fig:sim_results2} B.

\item The time required to complete the task is inversely proportional to the
  number of guides for the Hierarchical and the number of workers for the Egalitarian
  approach, as seen in Figure~\ref{fig:sim_scalability} and supplementary Figure~\ref{fig:egl_time}. This effect
  demonstrates that guide swarms achieve similar levels of parallelism as worker
  swarms, with the added advantage of effectively sharing information.
  The enhanced performance of the guides leads to higher success rates and
  shorter mission completion times, with fewer robots and reduced cost.

\item A small number of more proficient robots can better combine the
  information gathered. Despite larger coverage by Egalitarian swarms (shown in
  Figures~\ref{fig:sim_results2} D, F, H, and Figures~\ref{fig:model_diff} G-I), the lack of density hinders
  effective information sharing, primarily due to limited spatial awareness.
  Guides, while still requiring proximity to communicate, can identify
  previously explored areas and confirm target locations.

\item Heterogeneous exploration reduces mission success: workers often lack
  nearby robots to pass on findings and get stuck trying to go back to their
  initial location. Supplementary Figure~\ref{fig:first_target_detection} indicates that success might
  improve if, impractically, some workers could be excluded from exploration or
  could quickly return once a target is found.

\item Research indicates that the efficiency of robot swarms can significantly
  diminish beyond a certain number of robots due to congestion, whether physical
  or in communication~\cite{soma2023congestion,kuckling2023we,hamann2018swarm}.
  Implementing hierarchical structures within swarms can substantially reduce
  the number of robots required for completing tasks, effectively mitigating
  congestion-related issues at a cost that is comparable to, or even less than,
  that of an egalitarian swarm.
\end{compactenum}

Delivering multiple robots to a target requires global resource allocation for
specific tasks. Although this type of allocation may be possible with reactive
robots\cite{Ferrante2015EvolutionOS}, these improvements would likely increase
costs due to the swarm sensing area needing to match the environmental scale.

Drawing parallels to biological systems like ant colonies with distinct warrior
and worker roles\cite{simola2016epigenetic}, where different castes cover each
other's weaknesses, our setup sees guides providing spatial awareness and
workers offering radiation tolerance. However, our hierarchical system also
critically integrates \emph{decision-making}, particularly in directing workers
to targets. These factors lead us to believe that our primary finding is broadly
applicable: \emph{despite potential reductions in performance gaps, which come
  at the expense of design effort, a hierarchical swarm tends to be more
  cost-effective}.

The Hierarchical approach allows for multiple levels of decision making,
avoiding the pitfalls of a centralized system such as a single failure point or
information bottlenecks. Hierarchies preserve key properties like scalability
and efficiency through local interactions at each
level\cite{dorigo2020reflections}, and feature \emph{graceful degradation} and
\emph{progressive augmentation}\cite{li2019decentralized}.
% , by adding or
% removing levels of the hierarchy

We also show that our hierarchical system is indeed robust in real-world
deployments. Finally, the Poisson model could be used to
properly size a swarm for a given application, accounting for
cost and time to complete its task.

% \section*{Outlook}

% Our experiments, including simulations and real-world hardware
% deployments, have highlighted the advantages of hierarchical organization over
% egalitarian swarm approaches.
% Simulation results have shown that hierarchical
% systems enhance the real-world applicability of robot swarms. At the same time,
% our hardware experiments have validated the hierarchical approach by
% successfully executing a radiation cleanup mission, aligning closely with our
% simulation outcomes and lending greater significance to our findings. In
% essence, our work not only provides a promising approach for hierarchically
% organizing robot swarms but also demonstrates their potential to excel in
% practical missions, offering valuable insights for the broader field of
% distributed intelligent systems.
Further development of this work would be to study \emph{dynamic} hierarchies,
where robot roles depend on information content, task, and position. Additional
study could determine parallels with known social and biological hierarchies and
further our understanding of the conditions of their formation.

\section*{Materials and Methods}
\subsection*{Mission Model}
Each robot has a disk-shaped sensing radius of $s_r$, within which it can sense
the environment for targets. A point $x\in\mathbb{R}^{2}$ in the environment is
said to be covered by a robot if the point $x$ lies within the disk-shaped
sensing range of the robot $s_r$. This sensing model creates regions in the
environment that become covered or uncovered depending on the location of the
sensors. The covered regions are the union of the disk-shaped sensing of all the
robots. Assuming a uniform and independent initial robot distribution on a 2D
plane, this configuration of the stationary robots can be described by a Poisson
Boolean model $B(\lambda,s_{r})$, with intensity $\lambda$. Extending this model
to mobile robots creates instances of the Poisson Boolean model over time, and
the resulting process forms a Poisson point process with some intensity
$\lambda$. In our mission, the robots move randomly in a direction $\theta \in
[0,2\pi)$ with velocity $v_r$ that maximizes the explored area. When assuming a
random mobility model that maximizes coverage and with robots having to detect a
stationary target in the region, it forms a Poisson process of intensity
$\lambda=2 \rho s_r v_r$ as found in\cite{Liu2013dynamiccoverage}, with $\rho$
being robot density in the arena. Given a set of robots uniformly and
independently distributed on a plane moving according to a random mobility
model, the fraction of area covered at time t is
\begin{align*}
f_a(t)=1-e^{- \lambda \pi s_{r}^2 }, \quad \forall t \geq 0.
\end{align*}
The fraction of area covered during a time interval $[t_1,t_2)$ is
\begin{align*}
f_i(t_1,t_2) = 1-e^{-\lambda \mathbb{E}(\alpha(t_1,t_2))},
\end{align*}
where $\mathbb{E}(\alpha(t_1,t_2))$ denotes the expected area covered by a robot
during the time interval $[t_1,t_2)$.

Let $X$ be the arrival time of a robot at the target, the probability of a
target not being detected by the robots decreases exponentially over time:
\begin{align*}
P(X>t) = e^{-2 \rho s_r v_r t} 
\end{align*}

The expected arrival time at the target is $\mathbb{E}[X] = \frac{1}{2 \rho s_r
  v_r}$ with variance $Var[X] = \frac{1}{(2 \rho s_r v_r)^2 }$. The expected
arrival time is inversely proportional to the number of robots, sensing range and velocity of the robots. As discussed earlier, increasing the number of
robots seems to be an effective strategy to achieve the desired arrival time by
achieving the required collective sensing area. The guide robots used in this
study generate a map and can hence represent a target location in its map; this
enables the guide to transfer the target location to other guides or lead
workers to the target. With the availability of guide robots in the swarm, only
one guide robot must identify the target location, even with a mission
requirement requiring ten workers to reach the target location. Conversely, when
the swarm is only composed of worker robots, as in the egalitarian strategy, all
ten workers must reach the target by navigating the region. The arrival time
$T_n$ of n worker robots reaching a target forms an $Erlang(n,\lambda)$
distribution\cite{pishro2014introduction}, with $n$ being the number of
arrivals and $\lambda$ being the intensity. The Probability Density Function
(PDF) of n target arrivals takes the following form:
\begin{align*}
f_{T_{n}}(t)=\frac{\lambda^n t^{n-1} e^{-\lambda t} }{(n-1)!}, \quad \forall t > 0
\end{align*}
The expected arrival time of n worker robots at the target is $\mathbb{E}[T_n] =
\frac{n}{\lambda}$. Figure~\ref{fig:model_diff} A-C shows the difference between the
model-estimated mission completion times and the empirical outcome when using
the Egalitarian strategies, and figure~\ref{fig:model_diff} D-F shows the difference
with the Hierarchical strategy. %\squestion{Do we set n=10 for egalitarian
%  methods?}

\subsection*{Robot setup}
Our study employs a heterogeneous robot swarm comprising two distinct robot
types: guides and workers. Robots utilize software components based on their
hardware configuration, which are deployed using the Robot Operating System
(ROS)\cite{cousins2010sharing}, with slight variations between simulation and
hardware experiments. Specifically, in the simulated setup, the localization
module is replaced with a positioning sensor that provides pose estimates of the
robots, and the communication module is replaced with the range and bearing
sensor implemented in ARGoS3\cite{Pinciroli2012}.

\noindent Each guide robot is equipped with four key modules:
\begin{compactenum}
\item Perception and Mapping: This module includes a 2D lidar odometry
  package\cite{hess20162Dlidar}, generating a pose estimate (x, y, $\theta$),
  and a stereo visual fisheye camera (Intel T265), which produces a 3D pose
  estimate in $\mathbb{R}^3$ along with a rotation in SO(3). A Kalman filter
  fuses the pose estimates from both sensors to obtain a unified pose in
  $\mathbb{R}^3$ and rotation in SO(3). The mapping module employs a volumetric
  mapping system\cite{oleynikova2017voxblox} that incrementally constructs a
  Truncated Signed Distance Field (TSDF) using point cloud data. This TSDF is
  subsequently utilized for exploration planning.
\item Planning: The planning module encompasses two main planners:
\begin{inparaenum}
\item Exploration Planning\cite{dang2020graph}: This planner identifies the
  optimal path to maximize volumetric gain during exploration.
\item Local Planning\cite{macenski2020marathon2}: It ensures robot
  traversability and safety by planning local trajectories.
\end{inparaenum}
\item Behavioral Control and Communication: The behavioral control component
  encompasses the behavioral scripts that enable decision-making and
  communication within the swarm.
\end{compactenum}
We refer the reader to the supplementary materials for more details on software modules and their components.

\noindent The worker swarm consists of straightforward robots with a fisheye
stereo camera system and a Raspberry Pi4\cite{rpi4} as their computing
platform. Both worker and guide robots are tagged with visual markers to
identify neighbouring robots and their positions within the camera's field of
view. A tag detection mechanism\cite{olson2011tags}, specifically using
April tags\cite{april_github}, processes the fisheye images on the Raspberry
Pi4 to determine the positions of nearby robots in real time. The Khepera IV
base robot provides essential sensors, including 8 proximity sensors (range
0-25cm) and 5 ultrasound sensors (range 25cm-2m). Ultrasound sensors are
employed for obstacle detection and avoidance. 

% Figure 8
\begin{figure}[H]
\includegraphics[width=\textwidth]{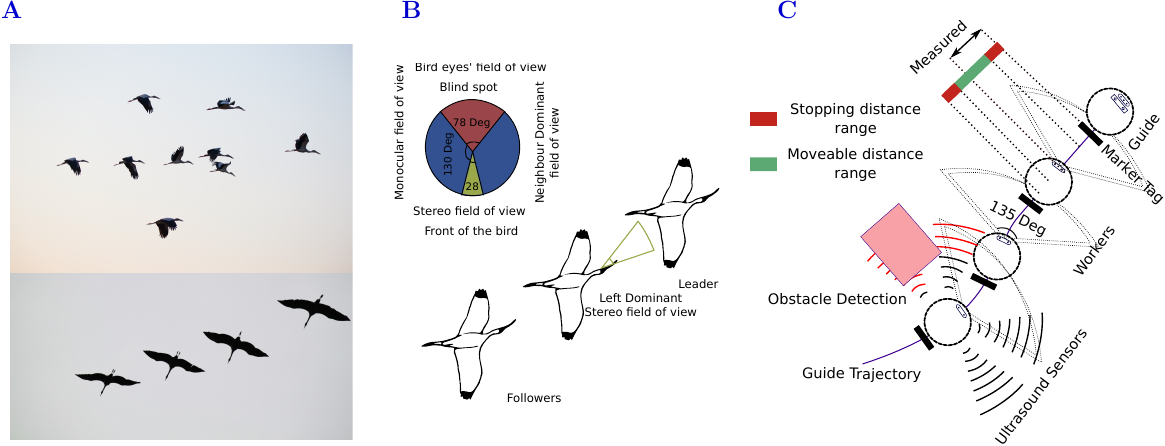}
\caption{Illustration of the chain formation behavior in comparison to white storks behavior: (A) a flock of white storks
  flocking during a migratory flight. (B) illustration of the migratory flocking
  performed by white storks by each aligning their front neighbour towards its left
  stereo field of view to exploit the thermal exploration performed by the
  leader bird in front. (C) the algorithmic behavior implemented on the guide
  robots and worker robots to allow exploitation of guides capabilities to
  navigate the environment.} 
\label{fig:chain_method}
\end{figure}

As shown in
equation~\ref{equ:harmonic_potential}, the potential function helps the robots
maintain a safe distance from obstacles by modifying their movement commands (see
Figure~\ref{fig:chain_method} C).
\begin{align}
  \label{equ:harmonic_potential}
  \phi(d) =
  \begin{cases}  
   a_0 + \frac{k d^3}{2\lvert d \rvert} & , d < d_t \\
   0 &, otherwise
  \end{cases}
\end{align}
with $a_0=0$, $k=0.2$ being design constants and $d_t=0.5m$ being a distance
threshold to maintain from the obstacles.

\noindent The Mobile Ad-hoc Network (MANET) \textbf{communication} system is
uniform across both guides and workers. Utilizing standard WiFi hardware
available in off-the-shelf computers, a MANET is established, enabling
peer-to-peer communication without the need for a router or additional radio
equipment. The MANET is created using BATMAN-adv\cite{batman}, which also
facilitates neighbour discovery. Communication within the robot swarm is based on
a gossip algorithm, employing local broadcasts via TCP/IP to communicate with
neighbouring robots identified through AVAHI IP resolution. To prevent network
flooding, a broadcast message quota of 500 bytes per control step (running at
10Hz) is imposed. Messages exceeding this quota are prioritized. Additionally,
guides exchange map layers (i.e., TSDF) with other guides. The modified TSDF layers' incremental differences are
periodically broadcast using the Data Distribution Service (DDS) with a map
exchange policy set to \textit{best effort} and \textit{volatile}.

\section*{Behavior setup}
The high-level behavior governing the robots evaluates specific conditions and
collectively defines a set of tasks for them to execute. The task allocation
behavior enables robots to select and engage in one or more of these tasks. Once
a robot takes on a particular task, it triggers a corresponding local behavior
to execute it. These local behaviors persist until one or more of their
associated local goals are accomplished, alone or collectively. This behavioral
framework is instrumental in facilitating the prototype radiation cleanup
mission.

We use Buzz\cite{pinciroli2016buzz}, a domain specific behavior scripting language for heterogeneous
robot swarms. Robots using Buzz contain a Buzz Virtual Machine (BVM) to execute
a script that describes the behavior of the robot. It is worth noting that all
robots (workers and guides alike) share the same script, like ants with the
ability to switch roles despite their morphological
differences\cite{simola2016epigenetic}. The script implements all the behaviors
described for the mission and only the behaviors that are appropriate for the
robot and the task are triggered based on the current conditions. The high-level
state machines described in Figure~\ref{fig:state_machine} J-K is a flattened state
machine depicting the sequence of action states taken by the robots in a given
type of control and some of their behavioral states by control strategy in Figures~\ref{fig:state_machine} A-I.

\subsection*{Radiation cleanup}
The Radiation cleanup behavior comprises a straightforward set of states:
Search, Identified, Cleanup, and Ideal. During the Search state, the robots are
tasked with exploration, a task assigned to all robots capable of exploring
based on the underlying control method. Transitioning into the Identified state
occurs when robots acquire knowledge that one of the robots has located a
target. While in the Identified state, the robots collaborate to determine which
ones will engage in the cleanup task through task allocation. Those committed to
cleanup then navigate to the target and commence the cleanup process. In both
the Identified and Cleanup states, if additional targets are available, robots
not committed to cleanup revert to the Search state to identify new targets.
When no more targets are present, these non-committed robots return to the Ideal state (i.e. Mission complete). This high-level behavioral sequence encapsulates the mission to ensure an
efficient and organized approach to radiation cleanup.

\subsection*{Task Allocation}
Task Allocation is a critical aspect of the robot swarm's coordination, and it
occurs in two key states during the radiation cleanup mission: Search and
Identified.
\begin{compactenum}
\item In the Search state, task allocation aims to determine whether a robot
  possesses the necessary skills to perform the exploration required by the
  selected control method (Heterogeneous, Hierarchical, or Egalitarian). If a robot is
  deemed suitable for exploration, it is assigned the task of exploring the
  environment.
\item During the Identified state, task allocation focuses on selecting the lead
  robot for chain formation and, if necessary, additional guide robots to join
  the chain. Once the lead robot is identified, the robots collaborate to form a
  chain that will guide them to the target location efficiently.
\end{compactenum}

\subsection*{Exploration and Homing}
Guide robots utilize a graph-based exploration planner
(GBplanner)\cite{dang2020graph} for exploration and homing. This planner
generates trajectories that maximize the volumetric gain of the map while
exploring. As guides explore, they construct a global graph of feasible states
within the environment. This graph, in conjunction with the Truncated Signed
Distance Field (TSDF) map, aids in exploration and navigation. The stored global
graph is used to calculate paths to various locations, including the home base.
A local planner refines these trajectories to ensure they meet traversability
requirements and are then translated into actuation commands for exploration.

In contrast, worker robots employ a bug-inspired exploration algorithm known as
the Swarm Gradient Bug Algorithm (SGBA)\cite{McGuire2019}. This algorithm
employs a simple state machine with four states and utilizes proximity sensors
to explore unknown environments. Workers rely on the gradient of the home
beacon's signals for homing, moving toward the source based on signal strength.
In the Heterogeneous control approach, workers and guide robots explore the environment
to identify targets. When a worker robot identifies a target, it acts as a
beacon to attract other workers, as it doesn't possess a map to communicate the
target's location. Workers within the field of view of another worker beacon
move towards it following the gradient--a behavior referred to as
\textbf{Neighbour Seek}.

\subsection*{Chain Formation}
The Chain Formation behavior is activated when the radiation cleanup mission
reaches the Identified state, and it becomes apparent that worker robots are
stationed at the home location, ready to be mobilized to the target. Worker
robots actively exploring the environment initiate the homing behavior as soon
as they receive word of target identification by a guide robot.

Once the lead guide robot responsible for guiding the chain to the target is
determined through task allocation, it initiates the chain formation process.
During chain formation, robots communicate via broadcast messages to establish
their position within the chain. Specifically, a turn-taking approach is
employed to arrange the robots in a linear chain formation, with each robot
positioned behind another.

Prior to the initiation of chain movement, each robot in the chain is aware of
its parent robot and, its child robot, if applicable. The number of robots
required for a chain is designed to be configurable, offering flexibility for
users to specify the necessary number of robots to effectively perform the
radiation cleanup mission.

The chain formation and movement behavior takes inspiration from the migratory flocking performed by white stocks (see Figure~\ref{fig:chain_method} A), where the followers align with the leader using their field-of-view as in Figure~\ref{fig:chain_method} B. Figure~\ref{fig:chain_method} C show the behavior performed by the robots during chain movement.

During the chain movement, the robots execute a behavior known as \textbf{Neighbour Seek/Lead}, which comprises three states:
\begin{compactenum}
\item Move: In this state, robots follow the relative position of their assigned
  parent robot, determined through tag detection and tracking (in simulation
  through range and bearing sensor). This state is active when both the parent
  and child robots are within an acceptable distance threshold, as illustrated
  in Figure~\ref{fig:chain_method} C. If a robot moves too far away
  from its parent, it broadcasts a stop command to request a pause from the
  parent. Similarly, if a robot gets too close to its parent, it stops moving
  and broadcasts a move request to the parent. The transition between the stop
  and move states for each pair of robots coordinates their motion and maintains
  the chain formation. This behavior is inspired by the coordination observed in
  white storks\cite{flack2018local}.

\item Wait: Robots in this state have received a stop broadcast from their
  parent and consequently stop moving, allowing the child robot to catch up. The
  wait state transitions back to the move state only upon receiving a move
  request from the parent, ensuring synchronized movement.

\item Recover: If a robot loses track of its parent's tag, it enters the
  recovery state. Recovery involves a 360-degree rotation motion aimed at
  regaining visual contact with the parent's tag within the robot's field of
  view. This recovery routine is particularly useful when robots need to make
  sharp turns ($>65$ degrees), potentially leading to the parent robot moving
  out of the field of view of the following robot. Additionally, in cases where
  intermediate robots are prone to failures, a simple broadcast within the chain
  of robots, as proposed in our previous work\cite{Varadharajan2020}, can
  bypass the failed robot and allow the mission to continue.

\end{compactenum}

%BibTeX users: After compilation, comment out the following two lines and paste in
% the generated .bbl file. 

\bibliography{scibib}

\bibliographystyle{Science}

\section*{Acknowledgments}
We express our gratitude to NSERC for funding this
research. Special thanks to the Digital Research Alliance of Canada for
providing essential computing resources for our experiments. We extend our
appreciation to the dedicated referees who diligently reviewed our paper and
provided invaluable feedback, including Dr. Vito Trianni, Prof. Marco Dorigo,
and Prof. Mauro Birattari. The authors also acknowledge the invaluable support
from members of the MIST lab, including Yanjun Cao, David St-Onge, Marcel
Kaufmann, Du Wenqiang and many others, who directly or indirectly contributed to this study.

\noindent\textbf{Funding:} The work was funded by NSERC Discovery Grant number 2019-05165.

\noindent \textbf{Authors' contribution:} VSV conceived the idea in collaboration with GB, developed and implemented the
algorithmic components, established the simulation environment, conducted data
analysis, generated plots, and played a key role in manuscript preparation. KS
made significant contributions to the simulations, data analysis, plot
generation, and manuscript preparation. SD was involved in the development and
implementation of algorithmic components and contributed to manuscript
preparation. PYL played a role in developing algorithmic components and
participated in manuscript preparation. GB substantially contributed to the
conceptualization of the idea, secured the study's funding, and actively
participated in manuscript preparation.

\noindent \textbf{Competing interests:} The authors declare that they have no competing interests.

\noindent \textbf{Data and materials availability:} All data needed to evaluate
the conclusions in the paper are present in the paper or the Supplementary
Materials. The source code and the data analysis scripts will be open sourced
under MIT license.

\clearpage

\section*{Supplementary materials}
Supplementary text section S1 to S4\\
Figures. A1 to A21\\
Tables 1 to 8\\
Videos 1 to 5\\
References \textit{(52-60)}

\renewcommand{\thefigure}{A\arabic{figure}}

\setcounter{figure}{0}

\textbf{\Large The supplementary PDF file includes:}
\begin{itemize}
  \item[] \large Section S1. Simulation setup
  \item[] \large Section S2. Hardware setup
  \item[] \large Section S3. Exploration, Localization and Coordination
  \item[] \large Section S4. Data analysis and statistics
  \item[] \large Figure A1. Map generated by the most updated guide robot with hierarchy and visitation frequency heat map of the worker robot when using egalitarian strategy.
  \item[] \large Figure A2. Normalised time to complete the mission when using an egalitarian strategy
  \item[] \large Figure A3. Average neighbour counts with an egalitarian strategy in all experimental configurations
  \item[] \large Figure A4. Effect of doubling the sensing range of the worker robots on the percentage of successful missions when using an egalitarian strategy
  \item[] \large Figure A5. Effect of providing 10x the experimental time limit on the percentage of mission success when using the egalitarian strategy
  \item[] \large Figure A6. The pose of 64 worker robots when using the egalitarian strategy at first target detection and mission completion
  \item[] \large Figure A7. Normalized time for the first robot to find the target using the three control strategies
  \item[] \large Figure A8. Hierarchical control state machine
  \item[] \large Figure A9. Egalitarian control state machine
    \item[] \large Figure A10. Heterogeneous control state machine
  \item[] \large Figure A11. Hardware specifications of the robot
   \item[] \large Figure A12. Guide robot software architecture
    \item[] \large Figure A13. Worker robot software architecture
    \item[] \large Figure A14. Illustration of the target placement in the environment
    \item[] \large Figure A15. Compound figure illustrating the behavioral hierarchy and types of robot available for deployment.
    \item[] \large Figure A16. Visual odometry and fused odometry of a guide robot during the experiments.
     \item[] \large Figure A17. Comparison of visual odometry to a motion capture system in a controlled environment.
     \item[] \large Figure A18. Mesh constructed using the D435 camera on the robots.
     \item[] \large Figure A19. Illustration of the coordinate system used on the guide robots.
     \item[] \large Figure A20. Behavioral hierarchy used by the robots.
      \item[] \large Figure A21. The structure of messages exchanged between the robots and the communication setup used on the robots. 
  \item[] \large Table 1. Cost of guide and worker robots
   \item[] \large Table 2. Power consumption of guide and worker robots
   \item[] \large Table 3. Data statistics on normalized time when using hierarchical control
   \item[] \large Table 4. Data statistics on explored area with hierarchical control
    \item[] \large Table 5. Data statistics on explored area with egalitarian control
    \item[] \large Table 6. Statistical test results on normalized time metric with hierarchical control
    \item[] \large Table 7. Statistical test results on explored area with hierarchical control
    \item[] \large Table 8. Statistical test results on explored area with egalitarian control
    \item[] \large Supplementary video captions
\end{itemize}

\newpage

\begin{figure}[H]
\includegraphics[width=\textwidth]{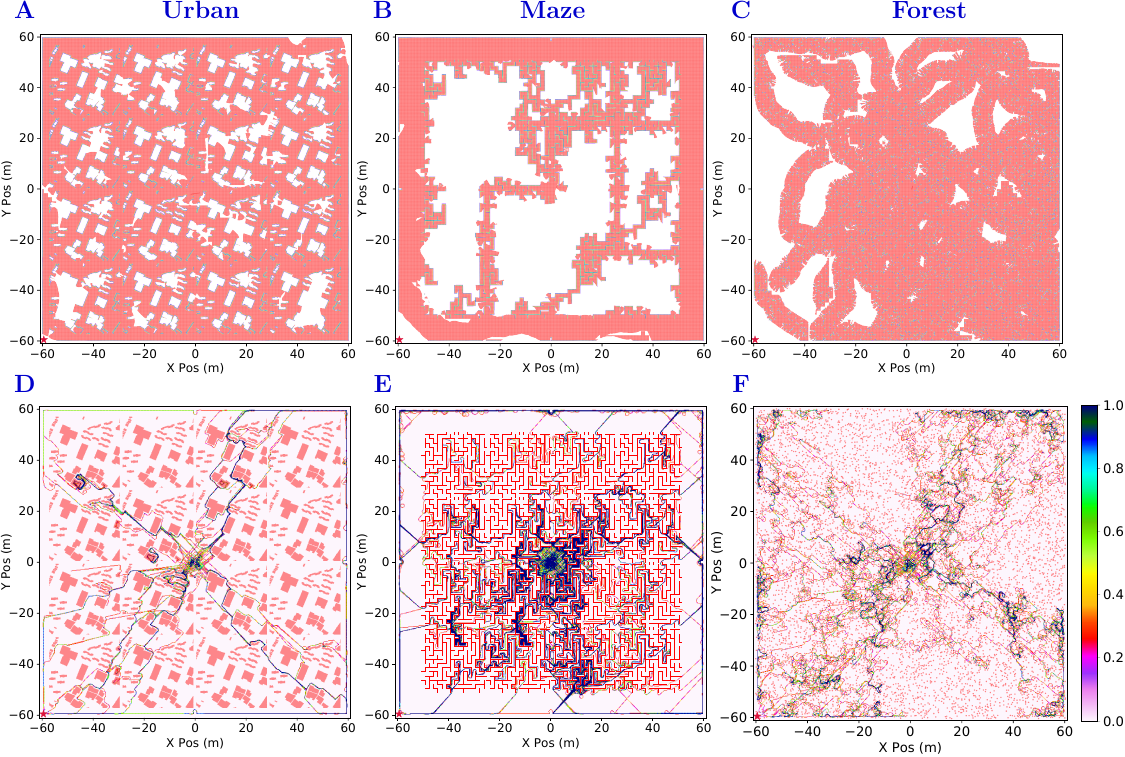}
\caption{The map generated by the most updated guide robot during one of the experimental run employing Hierarchical strategy in urban (A), maze (B), and forest (C). (D-F) show the visitation frequency heat map of the worker robot when using egalitarian strategy during three repetitions in urban (D), and maze (E), the forest (F) with single repetition as new forests are generated for each run.}
\label{fig:map}
\end{figure}

% Extended data fig 1
\begin{figure}[H]
\includegraphics[width=\textwidth]{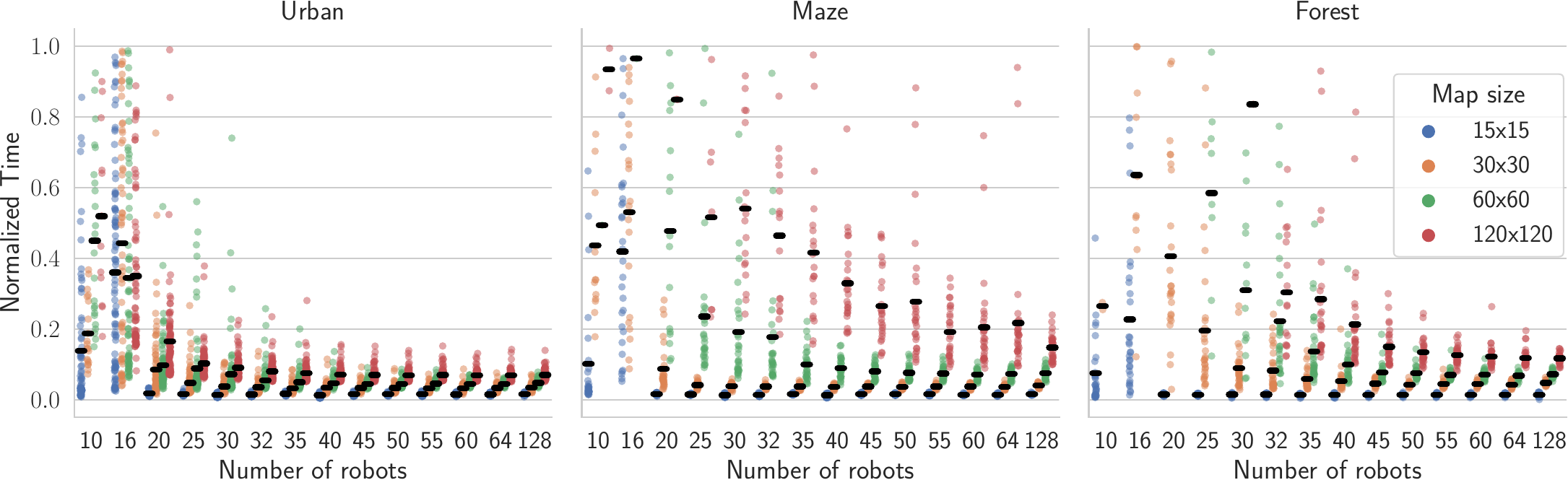}
\caption{The Normalised time to complete the mission in urban, maze, and forest environments with increasing number of robots when using an egalitarian strategy. The time taken to mobilize ten worker robots to the target using an egalitarian strategy exponentially decreases with more robots, in some cases, having no reported time as the mission was not successful. The exponential decrease in time is due to the increase in the collective sensing area of the swarm with more robots.} 
\label{fig:egl_time}
\end{figure}

%Extended data figures 

\begin{figure}[H]
\includegraphics[width=\textwidth]{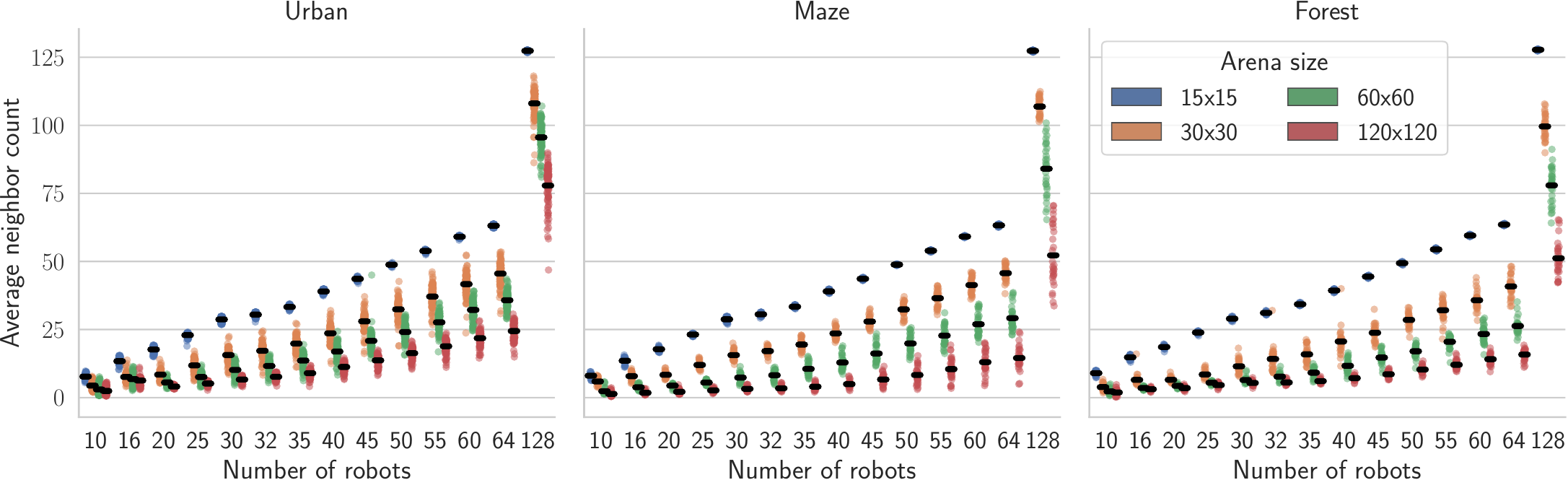}
\caption{The average neighbour counts with an egalitarian strategy in all experimental configurations. Intuitively, as the number of robots increases, the average neighbour count increases proportionally based on the size of the environment. The urban environment creates more neighbours than other arena types, as mission completion times are shorter to deploy all the robots. Maze and forest environments have a comparable number of neighbours, and when a sufficient number of neighbour counts is achieved, creating the required collective sensing area leads to mission success.} 
\label{fig:bug_nei_count}
\end{figure}

\begin{figure}[H]
\includegraphics[width=\textwidth]{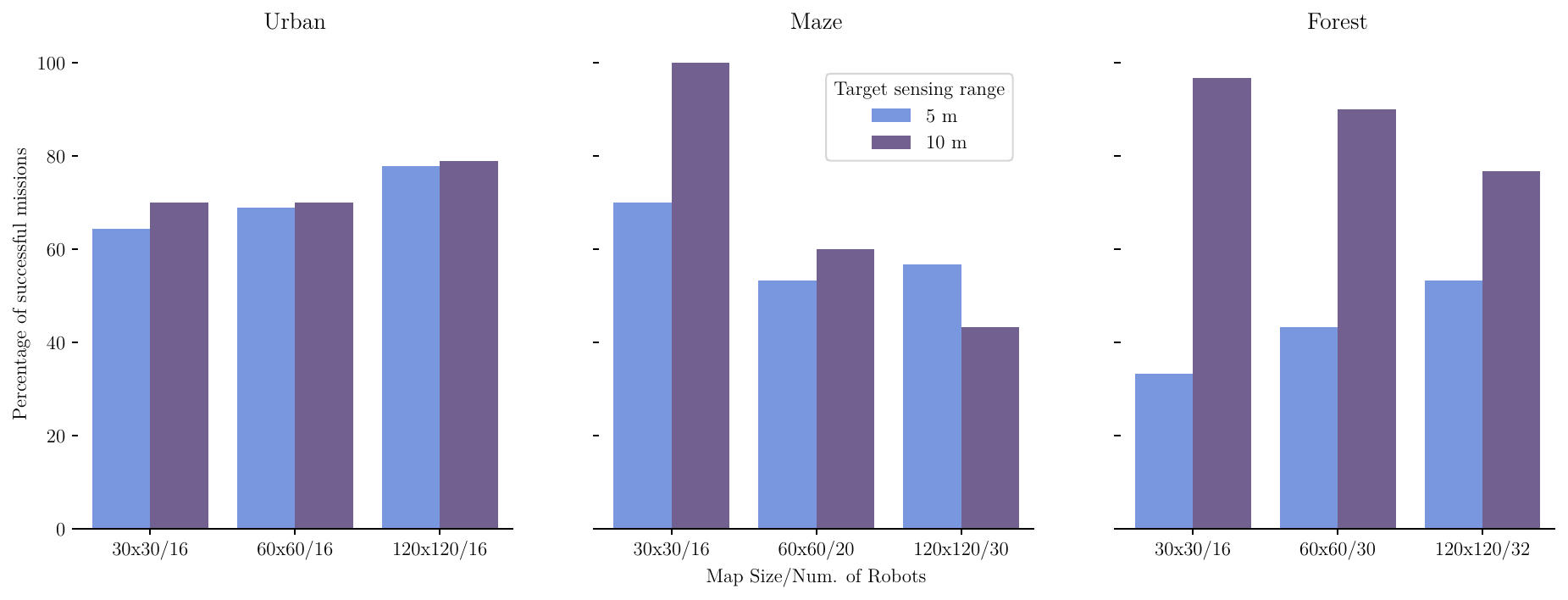}
\caption{The effect of doubling the sensing range of the worker robots on the percentage of successful missions when using an egalitarian strategy in urban, maze, and forest environments. The increased sensing range slightly increases the performance in urban and maze. The largest maze environments exhibit a drop in successful missions, where robots could get in a large loop revisiting already visited regions, indicating the need for sufficient collective sensing area achieved with an increased number of robots. The lack of structures in the forest environments favors the increased sensing range by exhibiting a significant increase in mission success.} 
\label{fig:bug_doubling_sensing_range}
\end{figure}

\begin{figure}[H]
\includegraphics[width=\textwidth]{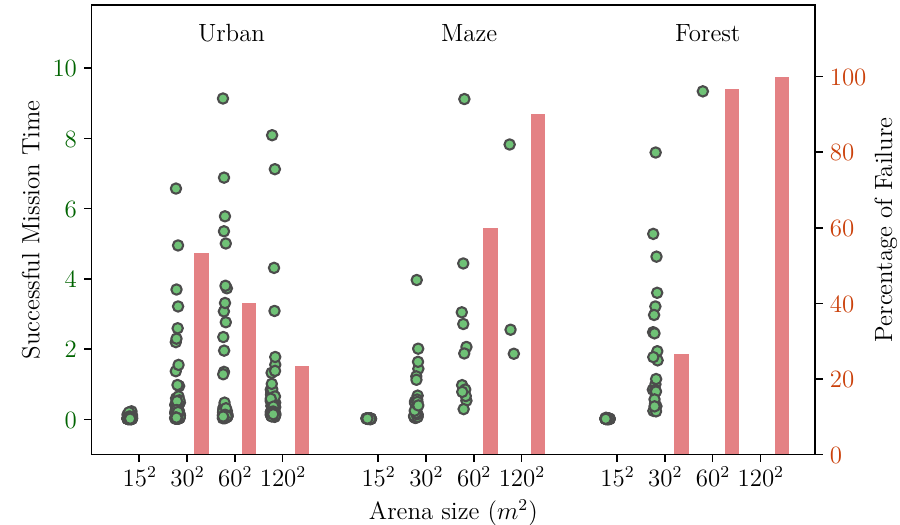}
\caption{The effect of providing 10x the experimental time limit on the percentage of mission success when using the egalitarian strategy with 16 workers in all three arena types. In urban environments, the percentage of successful missions increases with larger environments due to the increased presence of corridor-like free spaces induced by the tiling pattern of the environment creation. Straight corridors in urban environments enable the egalitarian robots to achieve a collective sensing area to enable mission success with sufficient time. The increased mission time does not improve mission success in maze and forest environments, as sufficient collective sensing could not be achieved.} 
\label{fig:bug_extended_time}
\end{figure}

\begin{figure}[H]
\includegraphics[width=\textwidth]{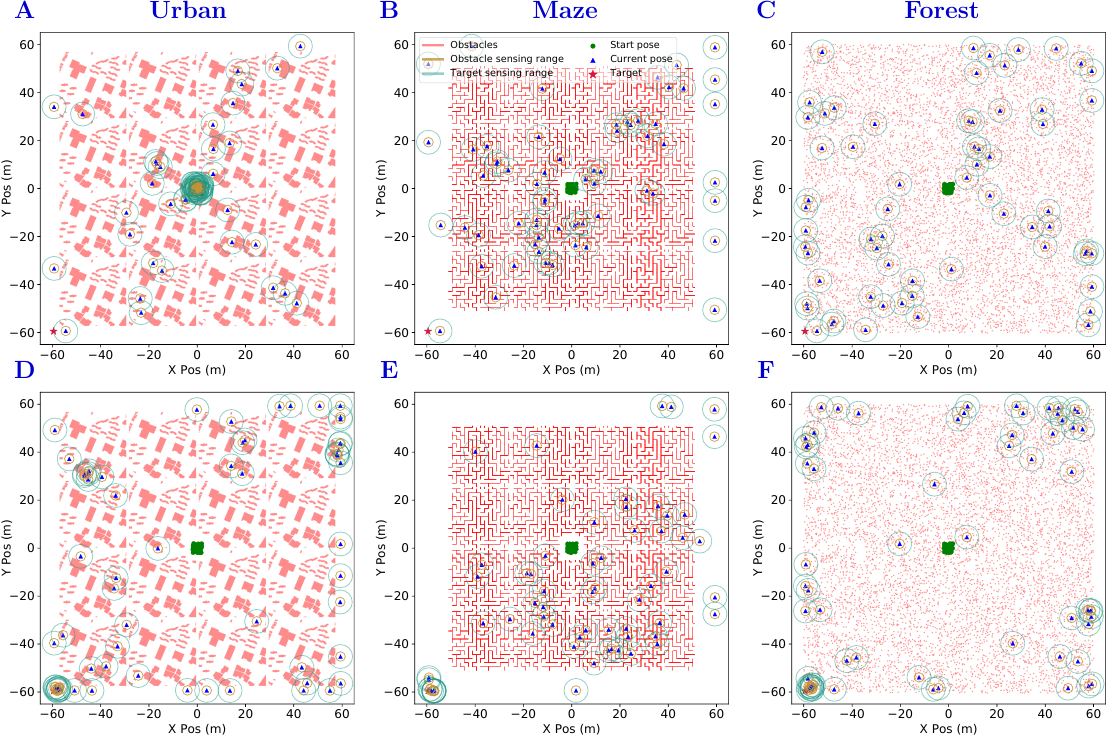}
\caption{The pose of 64 worker robots when using the egalitarian strategy: A-C shows the pose of the robots when a worker robot arrives at the target, and D-F shows the pose of the robots when all ten worker robots reach the target.} 
\label{fig:bug_pose_plots}
\end{figure}

\begin{figure}[H]
\includegraphics[width=\textwidth]{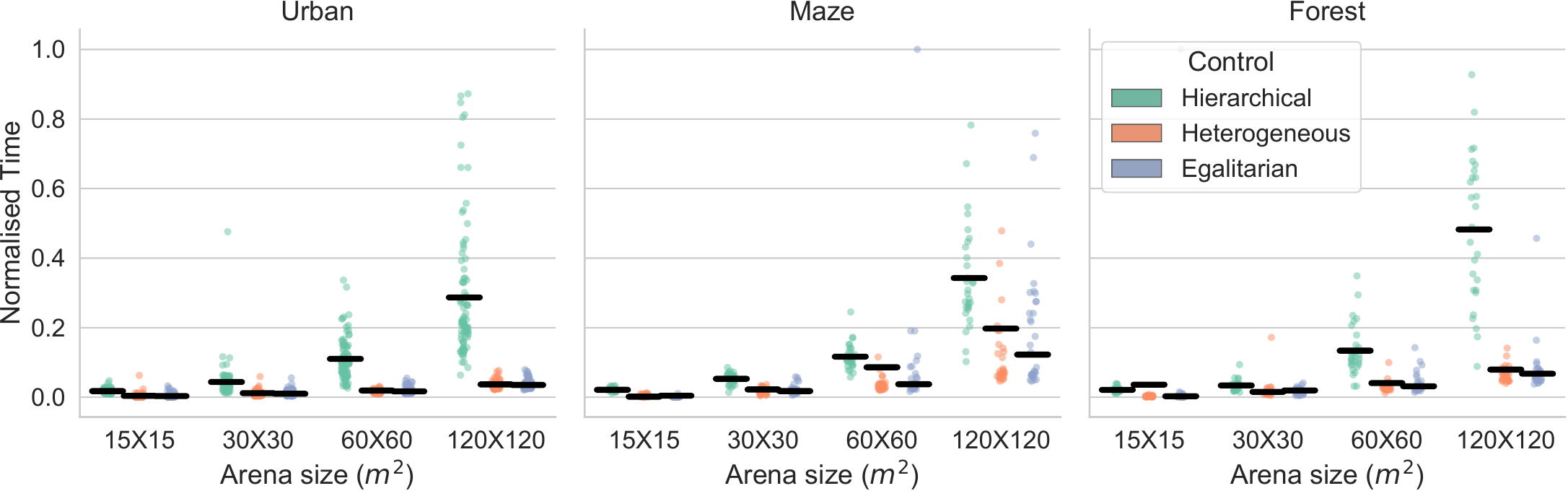}
\caption{The normalized time for the first robot to find the target using the three control strategies. The time for the first target detection stays higher for the hierarchical strategy than the other two as more robots explore in parallel. The higher performance in detecting the first target might not be beneficial when the swarm has to globally assign resources to a specific task, as in this work, arriving at the target with 10 workers.} 
\label{fig:first_target_detection}
\end{figure}

\newpage
\tableofcontents
\newpage
%\viveknote{Overall we are looking at something like the supplementary in the following link:}
% https://www.science.org/action/downloadSupplement?doi=10.1126%2Fscirobotics.abd8668&file=abd8668_sm.pdf

\section{Simulation setup}
The simulations were performed using the ARGoS3~\cite{Pinciroli2012} simulator, the simulator was choose because it enables simulation of large swarms of robots with physics and the modular structure enables addition of custom modules for environment creation and addition of custom sensors. The robot behaviors for the robots were designed using Buzz~\cite{pinciroli2016buzz}, a domain specific language for robot swarms. Buzz provides a set of preexisting programming primitives that simplifies the behavior design of multiple robots.
\subsection{Simulation software stack}
The software modules used by the robots were designed as ROS~\cite{quigley2009ros} nodes to leverage the advantage of modularity among the software components. The list of ROS nodes used during the experimental simulations:
\begin{itemize}
\item Control node: The control node implements a buzz controller with the ability to execute buzz scripts as a ROS node. The control node implements the robot behavior and connects to all the other software modules. The position estimates for the robots are obtained using the positioning sensor in the simulator and published to other nodes. The guide robots probe the exploration planning node to obtain an exploration and homing path during exploration. Worker robots interact with the gradient bug node to obtain the actuation commands during exploration. 
\item Mapping node: The mapping node implements Voxblox~\cite{oleynikova2017voxblox}, a volumetric mapping library based on Truncated Signed Distance Fields (TSDFs). The mapping node runs asynchronously and updates the map on reception of a new sensor frame. The library makes use of the 2D lidar and pose estimates to construct a volumetric representation of the environment. The volumetric map built by the module is used by the exploration node to compute exploration trajectories.
\item Exploration planning node: The software component implements the Graph-based exploration planner~\cite{dang2020graph}. The planner was interfaced with a ROS service that can be called to trigger planning and request an exploration path. The exploration path is received by the control node to execute the path through a publish-subscribe model.
\item Gradient bug node: The ROS node implements the swarm gradient bug algorithm~\cite{McGuire2019} (SGBA) as a ROS node. The node subscribes to the sensor readings and publishes an actuation command for the robot. 
\end{itemize}   

Each guide robot had a seperate unique control, mapping and exploration planning ROS node during the simulations. On the other hand, the worker robots had a seperate and unique control and Gradient bug ROS node.
\subsection{Control strategies}
The robot behaviors implemented on the robots enable the study of the three control strategy: Hierarchical, Egalitarian and Heterogeneous.
\subsubsection{Hierarchical}
In the Hierarchical strategy, the swarm is composed of both guide and worker robots. From a cost stand point, it is favorable to equip the swarm with a small number of guide robots, as they are three times the cost of a worker robots. In the uniform cost swarm comparison across different control strategies, two guide robots and ten worker robots were used.
\begin{figure}
\centering
\includegraphics[scale=0.7]{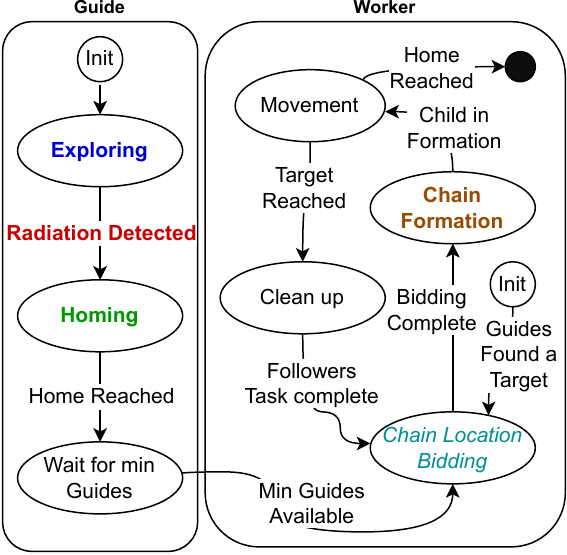}
\caption{The state machine illustrating the states taken by guide and worker robots during Hierarchical control.}
\label{fig:Hir_state_machine}
\end{figure} 
Figure ~\ref{fig:Hir_state_machine} shows the states taken by the robots during the mission. The guide robots perform frontier-based exploration using the exploration planner module in Explore State. The guide robots exchange the volumetric map with other guides in the communication range during exploration to produce a unified map. During the exploration, the guide robots detect the presence of a radiation source in the environment; the radiation source is simulated as a function of the distance to the target. On identification of the radiation source, the guide robots transition into the homing state. In homing, the control module requests the exploration planner for a path to the home location and navigates home. On reaching the home location, the guide robots wait for a required quorum size in wait for min guides state if more than one guide is needed to perform the worker mobilization. When the required number of guides are available, the guides transition to the Chain Location Bidding state. 

The worker robots initialize with the chain location bidding state, waiting for a guide robot to find and lead the worker to the target. The guide robots determine the robot that will lead all the robots to the targets. Ideally, the robot with the most updated map information leads the other robots. In the chain location bidding state, the guide leading the chain initiates the bidding by electing a worker robot that will follow the guide. This worker robot, in turn, elects another worker, and this process continues until the required number of workers is committed to the chain. All the other guides not committed to the chain take up a location at the end of the worker chain. The presence of additional guides could be used as a redundancy in case of a failure. On committing a required number of workers and guides to the chain, the robots transition to chain formation, where the robots committed to a location in the chain move and reach the location in the chain. On reaching a stable chain formation, the robots transition to the movement state, where the lead guide robot computes a path to the target and mobilizes the chain of robots to the target location.

\subsubsection{Egalitarian}
Egalitarian strategy makes use of only worker robots to explore and identify the target location in the environment. In the comparison of Egalitarian with other strategy, a uniform cost swarm across strategies with sixteen worker robots were used.    
\begin{figure}
\centering
\includegraphics[scale=0.7]{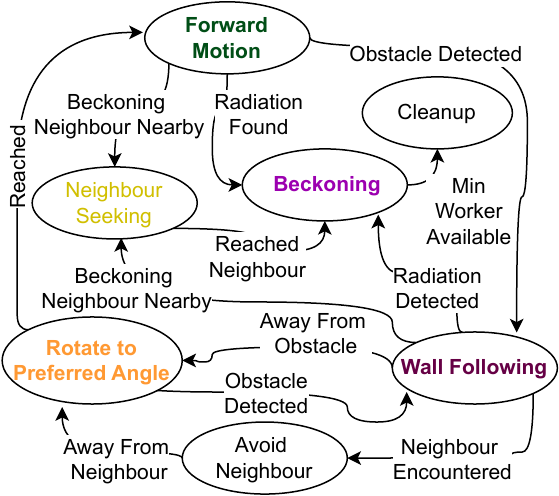}
\caption{The state machine illustrating the states taken by the worker robots with Egalitarian strategy.}
\label{fig:egl_state_machine}
\end{figure} 
Figure~\ref{fig:egl_state_machine} shows the states taken by the worker robots. The worker robots utilize the three primary states in SGBA: 1. forward motion, moving in a straight line when no obstacle is detected and when aligned with the preferred direction; 2. Wall following: when an obstacle is detected in the motion path, follow the contour of the object until the obstacle is not detected; 3. Rotate to the preferred angle when exiting the wall, following to continue moving in the preferred direction. During the execution of SGBA to perform exploration of the environment, the worker robots scan for the presence of radiation targets. If the robot detects radiation, the robot gets close to the source and starts beckoning to attract other worker robots within communication range. If another worker robot senses a beckoning robot, the robot approaches the beckon and attempts to find the target. With sufficient worker robots in a beckoning state, the robots perform radiation cleanup, and the mission is declared successful. The inability to store and communicate the target's location required each worker robot to reach the target on its own.

\subsubsection{Heterogeneous}
The Heterogeneous swarm was composed of 10 worker robots and 2 guide robots to create a uniform cost swarm and compare the performance across different strategies. 
\begin{figure}
\centering
\includegraphics[scale=0.7]{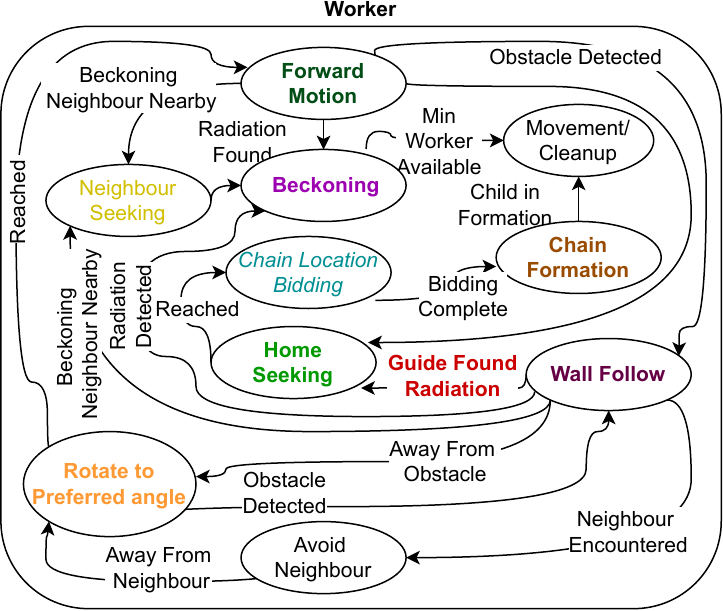}
\caption{The state machine illustrating the states taken by the worker robots with Heterogeneous strategy.}
\label{fig:hetro_state_machine}
\end{figure} 
With Heterogeneous strategy, both worker and guide robots explore the environment for targets using SGBA and frontier exploration, respectively. Figure~\ref{fig:hetro_state_machine} shows the states taken by the worker robot during Heterogeneous control, and the states taken by the guide robots are similar to Hierarchical control. The main difference with Heterogeneous control compared to other strategies is that all the robots use their potential to explore and rely on guide robots when the guide identifies a target. In particular, the worker robots explore using SGBA until a guide robot identifies the target. Once the worker receives a broadcast message indicating the identification of the target by a guide robot, it performs an inbound motion toward the home location. On reaching the home location, the worker robots wait for the guide robots to mobilize them to the target location using chain formation and movement. On reaching the home location with the target information, the guide robots wait for a predefined amount of time to arrive at the home location before initiating chain location bidding to determine the position in chain formation.

\subsection{Batch Simulation setup}
The experimental setup used in the study were designed to run multiple parallel simulations on compute clusters using a docker container. The desired experimental configuration could be passed as arguments to the experimental script that sets up the required experimental arena, targets, required number of robots by type, and configures the robots to use the required control strategy.
\subsubsection{Experimental configuration}
The required arena configuration and the data logging were achieved using the loop function within the ARGoS3 simulator. The loop function in the simulator can be designed as a separate module and plugged into a desired simulation. The Arena configurations used in the study are Urban, Maze, and Forest environments. Urban environments and mazes were obtained from a public database~\cite{sturtevant2012benchmarks} that provides the map in octile format. The required map file is parsed and loaded into the ARGoS3 environment. Forest environments were created by placing cylinders in the arena with a density of 0.1 and uniform distribution. This study used 30 unique urban maps, 10 unique mazes, and 30 unique forest maps with four different arena sizes (15X15m, 30X30m, 60X60m, and 120X120m). The various arena sizes were achieved using a tiling pattern that creates repetitive patterns of the same map file. Arena size of 30X30m represents the original size of the map file; for the 60x60m arena, the map file was tiled 4 times, and for the 120mX120m area, the map file was tiled 16 times. To achieve the smallest size, 15X15m, the map file was cropped to half the original size.

Each urban and maze environment was repeated three times in all experimental configurations. Creating 90 experimental runs in urban, 30 in maze, and 30 in forest environments for any configuration. The Egalitarian strategy was repeated with an increasing number of worker robots $N_W$.
\begin{align*}
N_W\in\{10,16,20,25,30,32,35,40,45,50,55,60,64,128\}
\end{align*}
\begin{sloppypar}
The Hierarchical strategy was studied with an increasing number of guide robots $N_G\in\{2,4,6,8,10\}$ and ten workers. To further examine the scalability with various targets, the hierarchical approach was studied with multiple numbers of targets $N_T\in\{2,4,8,16,32\}$ in the urban 60X60m arena. The Heterogeneous strategy was repeated for the uniform cost swarm of 2 guides and ten workers. A long experiment was repeated for the uniform cost swarm to study the effect of 10X the normalized time on the egalitarian strategy. With all the experimental configurations, a total of 12300 simulations were performed to obtain the statistics presented in the work.
\end{sloppypar}

\subsubsection{Target configuration}
The simulation of radiation targets in the simulation was using a target device that equips the communication module, the range and bearing sensor. The radiation simulated was a circular distance-based function that proxy for radiation propagation in the environment. In practice, a predetermined radiation threshold with a distance of five meters was used. A robot within the 5m disk-shaped region of the target triggers the presence of radiation. 

For all the experimental configurations, a single radiation target was placed in one of the corners of the environment, except for the Hierarchical approach scalability tests in the urban 60X60m arena. The placement of multiple targets was realized by symmetrically placing the targets on the border of the arena with different angular separations:
\begin{align*}
T_{\theta} = \{0,180,90,270,45,135,225,315,22.5,67.5,112.5,157.5,202.5,247.5,292.5,337.5\}
\end{align*}
from the center of the arena. For instance, the first two angular separations were used with two targets.

\begin{figure}
\centering
\includegraphics[scale=0.7]{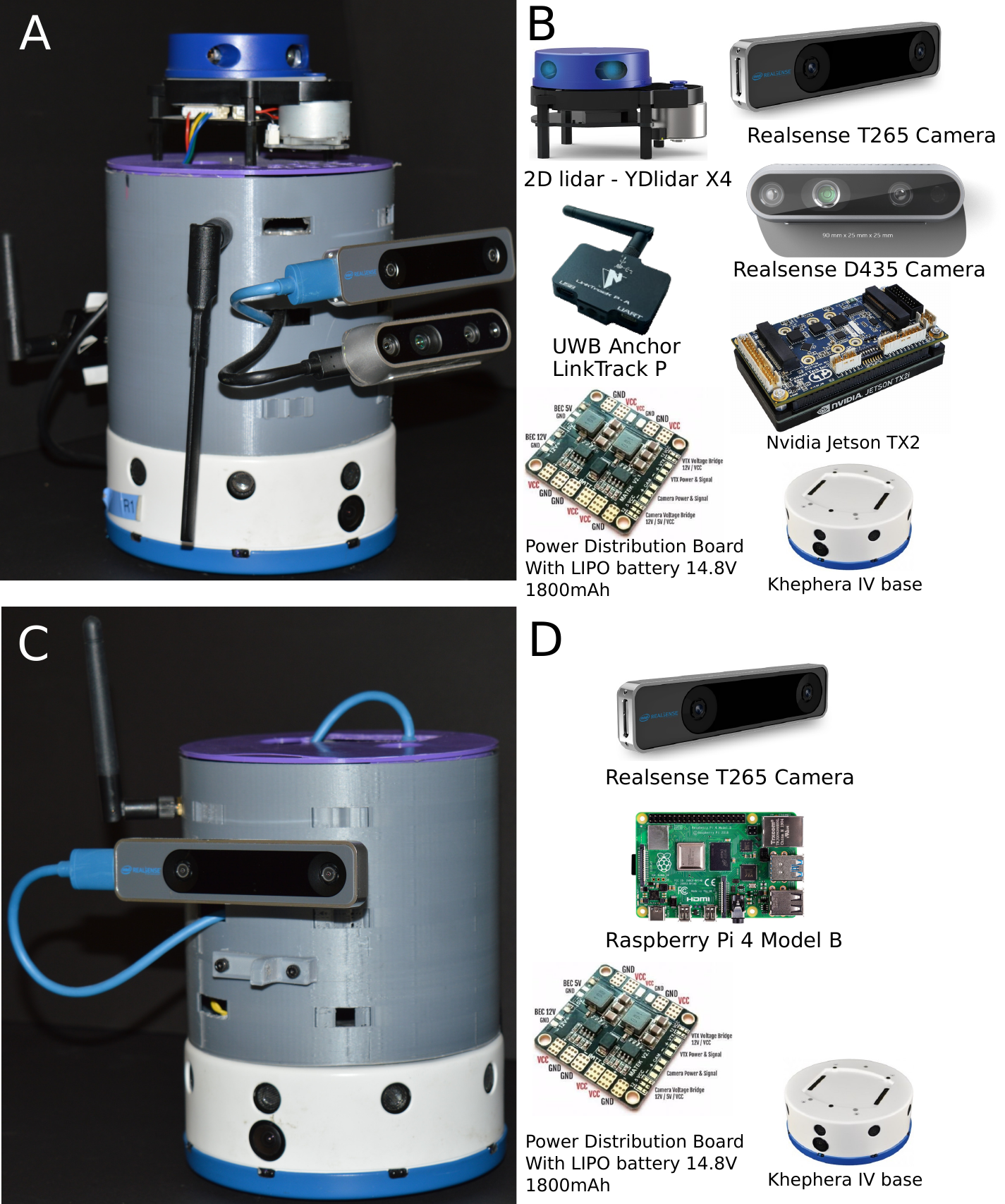}
\caption{Hardware specifications of the robots: (A) an image of the fully assembled guide robots called the super Khepera-IV robot, (B) show the sensing, computation and actuation components of the guide robot, (C) an image of the worker robot, and (D) components of the worker robot.}
\label{fig:hardware_specs}
\end{figure}  

\section{Hardware setup}
The hardware configuration detailed in this section enable the realization of the Hierarchical control strategy on the physical robots. 
\subsection{Hardware specifications}
The two robot types used during the experimental evaluations had different sets of sensors and computation platforms, as shown in figure~\ref{fig:hardware_specs}. The guide robot in figure~\ref{fig:hardware_specs} (A) was built with the components shown in figure~\ref{fig:hardware_specs} (B). The guide robot had a kheperaIV robot base with a 2D lidar, stereo fisheye camera module, stereo color camera module, a UWB anchor, and an Nvidia TX2 computer. The additional modules added to the top of the kheperaIV base were powered through a power distribution board and a 4S lithium polymer battery with a capacity of 1800 mAh.The additional sensors and the computer were hosted on the KheperaIV base using 3d printed parts and a casing. The worker robots (see figure~\ref{fig:hardware_specs} C-D) had a more straightforward setup, a 3d fisheye camera, and a Raspberry Pi4 computer on the KheperaIV robot base. Some worker robots had an Nvidia TX1 computer, demonstrating hardware-agnostic software with limited computational requirements. The worker robots used the same power distribution board and the battery as the guide robots to power the additional hardware.      

\begin{table}[htbp]
  \caption{Cost of the guide and worker robot.}\label{tab:robotcost}
\small
\centering
 \begin{tabular}{|c|c|c|} 
 \hline
% Header 
    Robot type & Component & Cost (\$) \\
 %  \multirow{2}{*}{Map type} & \multicolumn{5}{c|}{Size (Bytes)} \\
  % \cline{2-6} 
 %  &   ID & Key & Value & Extra & Total \\ 
 \hline
% Urban Entry 
\multirow{10}{*}{Guide}  & Khepera IV base & 3000 \\
 & Nvidia TX2 & 400 \\
 & carrier board - Elroy & 800 \\
 & Lidar & 120 \\
 & Intel Realsense D435 & 460 \\
 & Intel Realsense T265 & 350 \\
 & Power distribution Board & 15 \\
 & Battery & 50 \\
 & 3D print parts and screws & 30 \\
 \cline{2-3} 
 & Total & 5225 \\
\hline

\multirow{7}{*}{Worker}  & Khepera IV Base & 3000 \\
 & Power distribution Board & 15 \\
 & Battery & 50 \\
 & 3D print parts and screws & 20 \\
 & Raspberry PI4 B & 80 \\
 & Intel realsesnse T265 & 350 \\
 \cline{2-3} 
  & Total & 3515 \\
  \cline{2-3}
  & Cost ratio Guide to Worker & 1.486 \\
\hline
 \end{tabular}
\end{table}
The cost of the components used by the guide and worker robots are shown in table~\ref{tab:robotcost}. The main cost component in the hardware is the KheperaIV robot base, which can be easily replaced with a low-cost base. Apart from the base, the additional hardware components on the guide robot cost 2225 \$, and the worker robot costs 515 \$. The hardware cost ratio of the components with guide and worker robots amounts to 1.5211 a.u.. The guide robot costs 1.486 times the worker robot.

The power requirements of the guide and worker robot are shown in table~\ref{tab:power}. The overall power consumption of the guide robots is 23.6W, and the worker robot consumes 8.5W. The 4S Lipo battery with a capacity of 1800 mAh provides 1.12 hours of autonomy for guide robots and 3.1341 hours for worker robots. The power ratio of the guide to worker robots is 2.7764 a.u.. Considering the robots' hardware cost, power requirements, and autonomy time, we assign a threefold cost to the guide robots to compare results across different strategies with the two robot types.

\begin{table}[htbp]
  \caption{Power consumption of guide and worker robot.}\label{tab:power}
\small
\centering
 \begin{tabular}{|c|c|c|} 
 \hline
% Header 
    Robot type & Component & Power (W) \\
 %  \multirow{2}{*}{Map type} & \multicolumn{5}{c|}{Size (Bytes)} \\
  % \cline{2-6} 
 %  &   ID & Key & Value & Extra & Total \\ 
 \hline
% Urban Entry 
\multirow{7}{*}{Guide}  & Nvidia TX2 & 10 \\
 & Carrier board - Elroy & 7 \\
 & Lidar & 2.5 \\
 & Intel Realsense D435 & 2.6 \\
 & Intel Realsense T265 & 1.5 \\
 \cline{2-3} 
 & Total & 23.6 \\
 \cline{2-3} 
 & Autonomy time & 1.1288 hours \\
 
\hline

\multirow{5}{*}{Worker} & Raspberry PI4 B & 7 \\
 & Intel realsesnse T265 & 1.5 \\
 \cline{2-3} 
  & Total & 8.5 \\
  \cline{2-3}
  & Autonomy time & 3.13411 hours \\
  \cline{2-3}
  & Power ratio Guide to Worker & 2.7764 \\
  
\hline
 \end{tabular}
\end{table}

\subsection{Software Stack}
As in the simulations, the software modules used by the robots were designed as ROS nodes. Each robot had its own set of software modules to reason, plan, and actuate in the environment. A broadcast-based communication was used with the setup to propagate information among the robots in the swarm, as detailed in Section~\ref{sec:comms}. The list of ROS nodes used on the robot hardware: 
\begin{itemize}
\item Control node: The buzz controller node executes a buzz script implementing the high-level state machine that coordinates the information from all the other modules. The node steps at 10Hz to perform sensing, control, and actions. 
\item Mapping node: The node implements Voxblox, running asynchronously to produce a volumetric environment representation from the pose estimates and the Lidar. The node steps based on the rate of the sensor information; the Lidar produced ten frames per second, and hence, the node runs at 10 Hz. 
\item Exploration planner node: Implementing the Gbplanner for exploration, the node uses the volumetric map of the environment and poses estimates to produce an exploration trajectory. The node operates on a trigger from the control node for exploration and homing. The path produced during a trigger is sent to the control node for execution.
\item Lidar SLAM node: The node implements cartographer~\cite{hess20162Dlidar}, a 2d lidar Simultanious Localization and Mapping technique. The node provides the pose estimate using the odometry from the Lidar and provides a Lidar map of the environment. The node also performs map optimization to fix mapping errors. The node runs at the rate of 10 Hz.  
\item Kalman filter node: The node contains an instance of the robot localization~\cite{MooreStouchKeneralizedEkf2014} ROS node, a generalized Kalman filter implementation. The node obtains the pose estimate from visual SLAM and Lidar SLAM and fuses it to obtain more reliable odometry. 
\item Local Planner node: The local planner node within the ROS navigation stack. The planner uses a global and local cost map to navigate the environment safely. The planner uses the sensor measurements to build a cost map and moves the robot in a desired path by interacting with the control node.
\item Communication node: The node implements a UDP-based communication to broadcast information to all neighboring robots within the communication range. The node uses the Batman-adv neighbors function to determine the neighboring robots. The node obtains a serialized payload from the control node and broadcasts it to all the neighbors. 
\item Robot Driver node: The robot driver node interacts with the robot base through the UART serial port to obtain proximity sensor information and send actuation commands.
\item Sensor Driver nodes: A set of driver ROS nodes was used to obtain information from each sensor. A realsense ROS driver node for T265 and D435 camera, a lidar driver ROS node, and an UWB driver node.  
\item Tag Detection node: The tag detection node is used to undistort the fisheye camera images and detect tags within the image. The node provides the tag ID and the position of the tags to the control node to determine the neighboring robots' pose.   
\end{itemize}  

\begin{figure}
\centering
\includegraphics[scale=0.7]{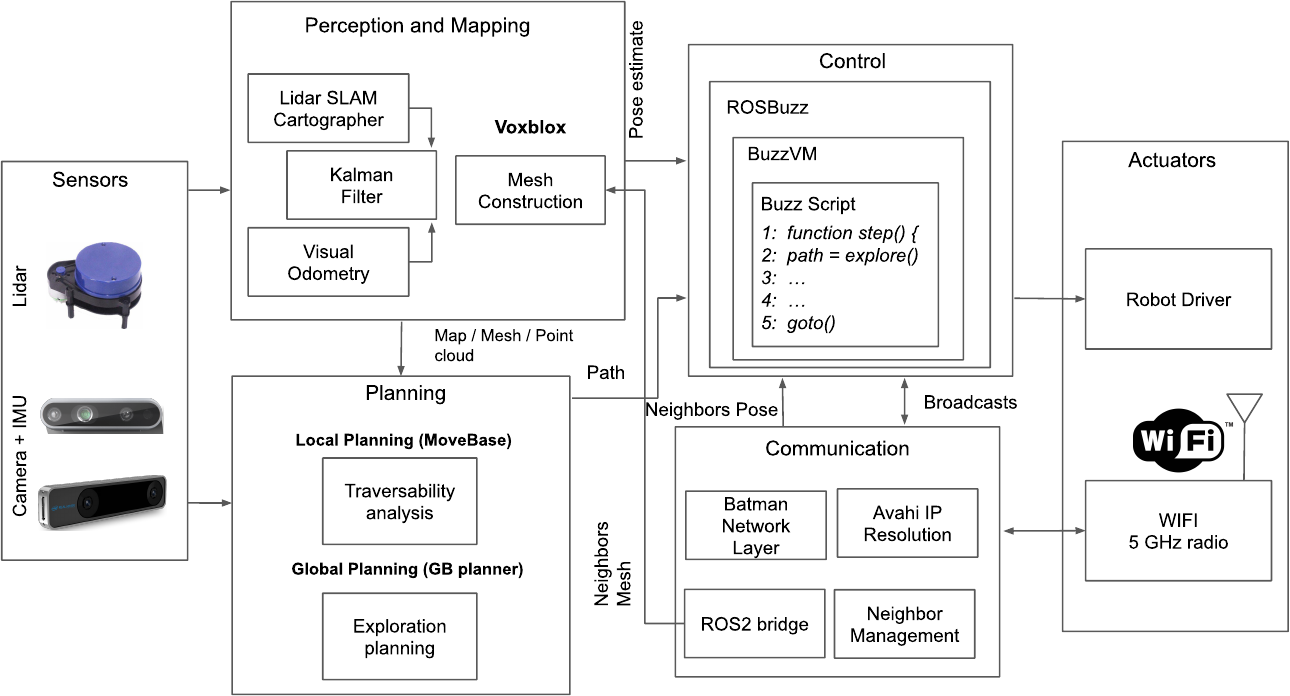}
\caption{Guides software architecture}
\label{fig:guidearc}
\end{figure}

\begin{figure}
\centering
\includegraphics[scale=0.7]{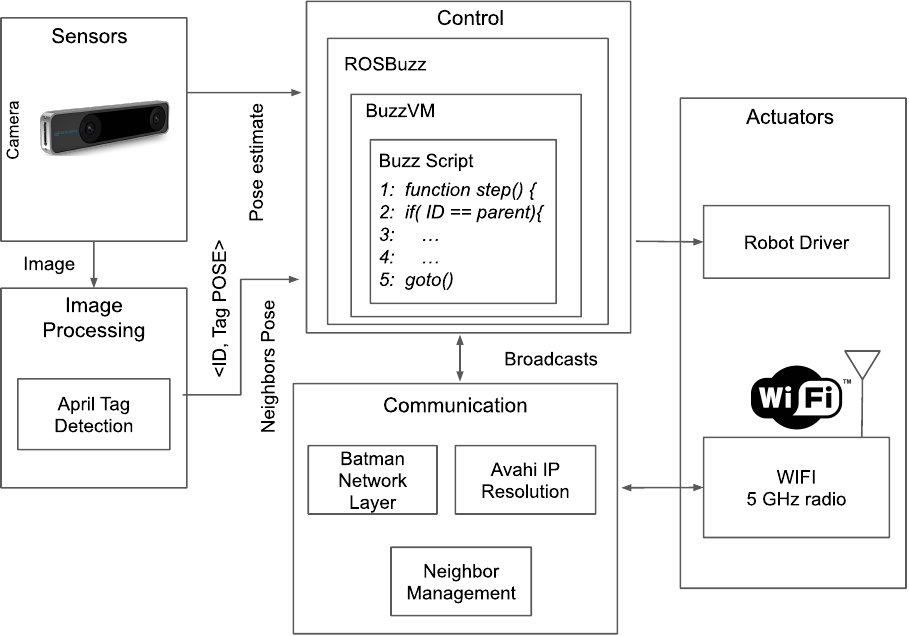}
\caption{Worker software architecture}
\label{fig:workerarc}
\end{figure}

Figure~\ref{fig:guidearc} shows the software modules on the Guide robots. The Guides had Control, Mapping, Exploration planning, Lidar SLAM, Kalman filter, Local planning, Communication, Tag Detection, and Robot driver ROS nodes. The information from all the nodes was used in the control node to realize the desired robot behavior. Figure~\ref{fig:workerarc} shows the software modules on the worker robots. The worker robots had Control, Communication, Tag Detection, and Robot driver ROS nodes. Compared to the guide robots, the worker robots had a simple set of software modules to realize the Hierarchical control strategy on the robots.

\subsection{Target representation}
\begin{figure}
\centering
\includegraphics[scale=0.7]{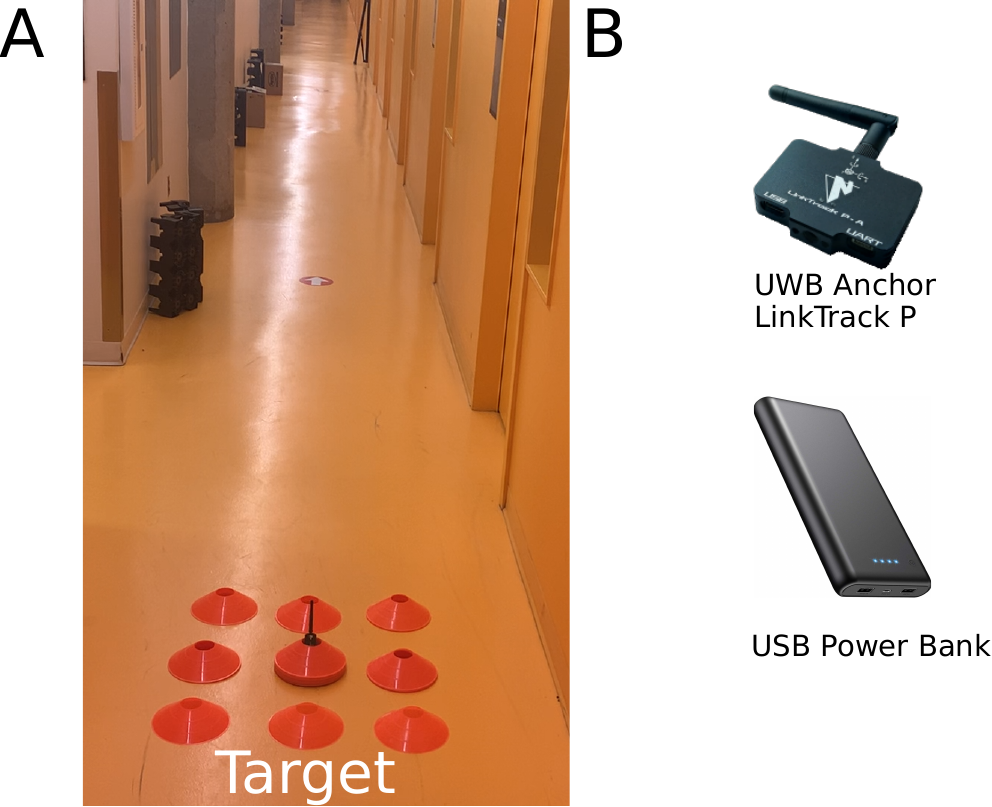}
\caption{(A) The placement of UWB tag in the corridor environment to represent the target. (B) The components of the target are an UWB tag and a battery bank to power the setup.}
\label{fig:target}
\end{figure}

The radiation targets in the environment were simulated using an Ultra-WideBand (UWB) tag as shown in figure~\ref{fig:target}. The guide robots and the target device had a similar tag from NoopLoop called LinkTrack P. All the devices were configured with an identical configuration to operate in \textit{DR mode} as tags with a unique ID. The nodes provide the following information of all the nodes in the range: 1. first path signal strength indication \textit{fp\_rssi}, 2. the total received signal strength indication \textit{rx\_rssi}, and 3. the estimated distance from Received Signal Strength Indicator (RSSI.) The node using the \textit{rx\_rssi} and \textit{fp\_rssi} can determine if a node is in Line Of Sight (LOS). When the target node is in LOS to the guide robot exploring, we use the distance estimated from RSSI to be less than the 5m sensing range of the robot to trigger the presence of radiation.

\subsection{Experimental Protocol}
The software on the robot is developed on the developer's computer using simulation and pushed into GIT repositories. The repositories for the robot are organized using the wstool tool, a command-line tool for maintaining a collection of ROS packages using a single `.rosinstall` file. The following is the sequence of steps that are performed for setting up the robot, running the experiment, and collecting the data:
\begin{enumerate}
\item Update code: The robot code is updated from the operator's computer using a bash script that runs the `wstool update` command on each robot.
\item Building software: The software is built on each robot to have the most updated software.
\item Starting the Camera: The four cameras placed in the environment are started to record the experiment.
\item Starting the Operator station: The operator computer was started with the required interface to visualize the robot's progress. The operator's computer was a standard laptop with communication hardware similar to the robots' (WiFi 5 GHz radio).
\item Launching the software stack: The software on the robots was launched using a roslaunch file with the required ROS nodes.
\item Triggering Exploration: The guide robots were triggered from the operator station to explore the environment and perform the mission.
\item Data collection: The logged data was transferred to the operator's computer using a bash script. The robots were logging a ros bag file with the map, position estimates, and sensor information and a Comma-separated value (CSV) file of the internal state of the robots from the control node. The ros bag and CSV files are transferred to the operator's computer.     
\end{enumerate}

\section{Exploration, Localization and Coordination}

\subsection{Guide robot exploration}
To explore the environment, the guide robots used a modified version of the GBPlanner~\cite{dang2020graph} from a team of the DARPA Subterranean Challenge~\cite{kulkarni2021autonomous}.
The exploration planner is divided into two parts: local and global planning.

The exploration local planner first extracts the bounds of the exploration space based on the local geometry of the environment.
The environment geometry is captured in the 3D mesh volumetric representation computed using the VoxBlox mapping technique~\cite{oleynikova2017voxblox}.
A sparse point cloud is obtained from the local volumetric map, and a bounding box of navigable space is extracted using the eigenvectors derived from its Principal Component Analysis.
Once the bounding box is established, the local planner performs traversability analysis by sampling points in the corresponding free space and connecting them with collision-free edges. 
The following step computes the volumetric gain associated with each vertices of the resulting 3D undirected graph. 
The volumetric gain corresponds to the expected amount of 3D information that is likely to be visible from a particular viewpoint. 
Then, the shortest paths to the highest volumetric gain vertices are obtained with Dijkstra's algorithm.
The path with the highest information gain overall is selected and refined for safety before being executed by the robot.
Safety refinement is performed using the ROS Navigation software stack~\cite{macenski2020marathon2}.

When local exploration is exhausted, or when the robot needs to come back to its home location, the global exploration planner is used to compute paths that exceed the sensors' range around the robot. 
The global planner maintains a sparse graph of the best local exploration paths in addition to sparsely sampled free-space points.
This information is sufficient to chart a course back to the home location as well as to select the most promising frontier to explore.
Frontiers of the explored space are leaf vertices in the global planner graph that have high volumetric gain.
Thus, the global planner is used to reposition the robot to one of these frontiers when local exploration is no longer able to yield significant gains. 
The frontier selection is based on multiple factors: volumetric gain, distance from current location, and the distance to the home location.
When the local exploration is again exhausted, the process is repeated, and the robot repositions itself to a new frontier.
In our experiments, the homing path to reach the starting location is computed once the robot has found the target.

In order to perform coordinated exploration with more than one robot, the guide robots periodically share their volumetric maps with one another, and, using the shared map representation of the environment, they perform exploration individually.
To improve coverage of the environment, the robots assign themselves, at the beginning, each a different preferred direction, which influences the frontier selection during global planning. 
This enables multiple robots to choose different sides at corridor intersections and avoid overlapping exploration paths. 

The map sharing was enabled between the robots by using a ROS1/2 bridge~\cite{ros2_bridge}, a ROS node that enables converting ROS topics to ROS2 topics. The conversion to ROS2 messages enabled the transfer of ROS topics over the mesh network to other robots. ROS2 used the Data
Distribution Service (DDS)~\cite{Hartanto2014} with a policy for map exchange set to \textit{best effort} and \textit{volatile} to send the map fragments to other robots. Periodically, the guide robots create messages with map fragments containing the incremental difference of the TSDF layers from Voxblox to be sent to other robots. The guides receiving the map message apply it to its current map with the transformation calibrated at initialization, merging the map of the two robots. Techniques like queue management with a predefined quota for each robot~\cite{saboia2022achord} might be helpful in congested networks requiring the transfer of large payloads. Since our exploration was not in the scale of kilometers, the size of the payload was managed with simple queue management from ROS.

\subsection{Navigation}
The guide robots use the local planner to navigate the environment by building a cost map. When a navigation path is obtained in the control node, the node parses the path points, sends an action goal, and waits for feedback. If a particular point cannot be reached, the control node, using the feedback, attempts to request the global planner for an updated path. During all three states' exploration, homing, and worker mobilization to the target, the guide robot uses the local planner to navigate the robot to its goal.

\begin{figure}
\centering
\includegraphics[width=\textwidth]{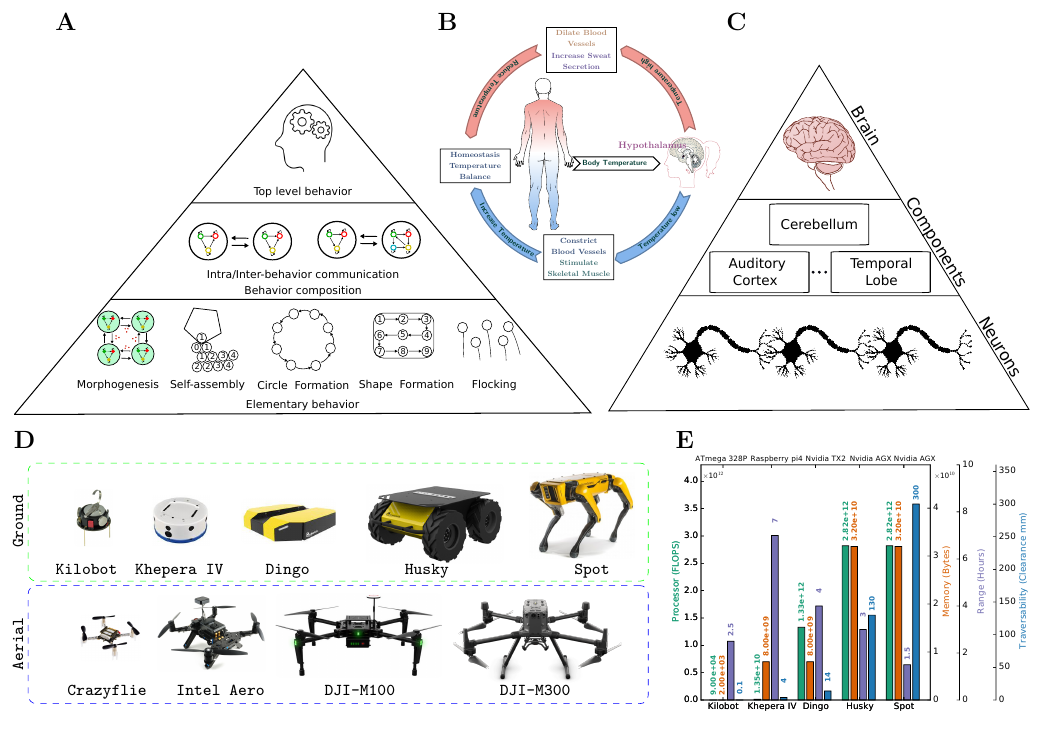}
\caption{Hierarchies in robot swarms. (A) primitive robots on the lower level
  perform collective behavior providing information and exploiting the distilled
  information coming from the higher levels, producing a synergistic swarm
  operation as a whole. (B) homeostasis is one of the tasks performed by the
  hypothalamus inside the brain: maintaining an optimal body temperature by
  sending stimulus to muscles, sweat glands and blood vessels. (C) The hierarchy
  represented by the organs in human biology, with the brain as a whole
  performing a complex task (maintaining the well-being of the individual),
  composed of a variety of components performing different sub-tasks, and down
  to a microscopic scale these components are made of more or less self-similar
  neurons. (D) a range of commercial-off-the-shelf robots that can be composed
  into a single swarm with hierarchies. (E) capabilities (processing capacity,
  memory, traversability clearance, range) for the ground robots shown in (D).
}
\label{fig:intro} 
\end{figure}

\begin{figure}
\centering
\input{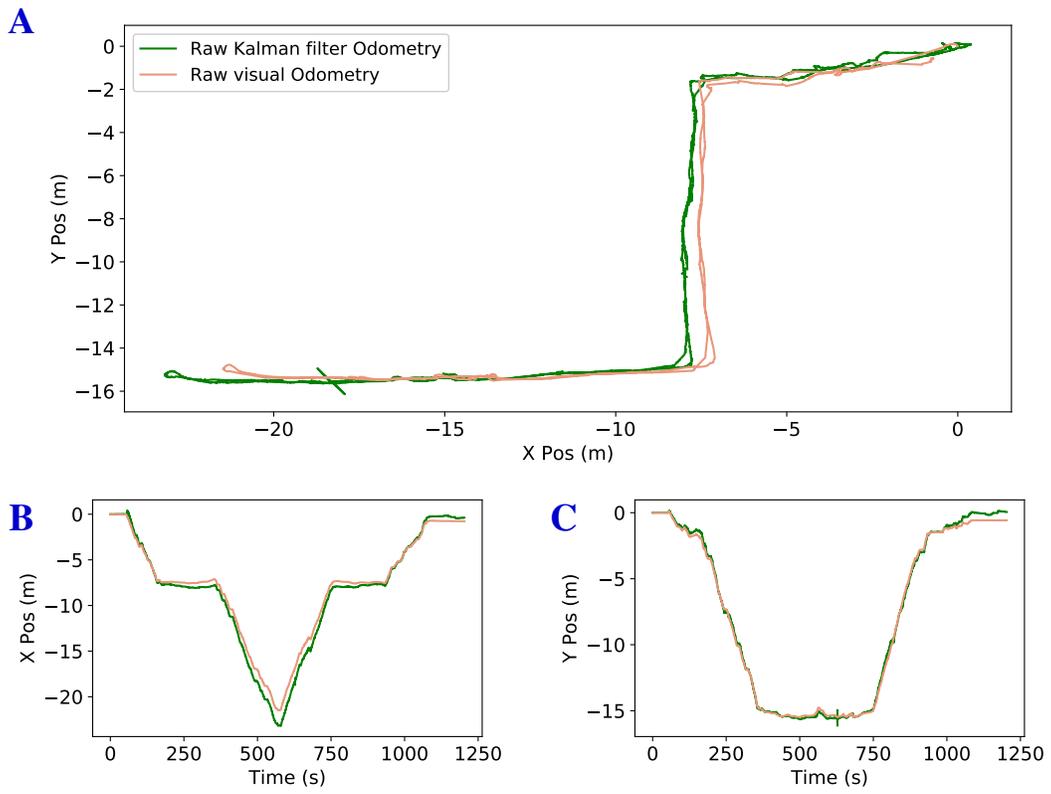}
\caption{(A) Visual odometry and Kalman filter estimate of a guide robot during exploration and homing. (B) Evolution of the X-axis pose estimate of the robot over time with visual and kalman filter fused estimates. (C) The evolution of Y-axis pose estimate of the robot.}
\label{fig:odom_diff} 
\end{figure}

\subsection{Positioning with and without SLAM}

\subsubsection{Localization for hierarchies}

A reliable localization for each robot in the swarm is a bottleneck in realizing an egalitarian swarm where all the robots can localize and map. A swarm where all the robots can localize might not be feasible using simpler sensors or non-reliable for the mission. Consider the corridor-like environment; there are several tight spots that can not be safely navigated without reliable odometry; mainly, the doorways are very challenging to navigate without reliable odometry. For instance, the shortest width of the experimental arena in this work was around 0.7m. A localization error of 0.3m is sufficient for the robots to miss a doorway. The localization accuracy of the robots depends highly on the type of sensors used; high precision Inertial Measurement Unit (IMU) and multi-beam Lidar could increase the accuracy by increasing the cost of the robots. Using the appropriate sensors and robots for a given type of mission minimizes the cost of equipment and operation cost. Minimizing cost with the ability to scale with the mission's complexity is one of the main advantages of hierarchies compared to egalitarian swarms.

Fig.~\ref{fig:intro} (A) shows the microscopic low-level behaviors that interact with a macroscopic high-level behavior through inter-behavior composition. This architecture is comparable to that of human physiology (see fig.~\ref{fig:intro}-C), with one striking example being human homeostasis, maintaining body temperature in humans through hierarchical control. Fig.~\ref{fig:intro} (D) shows a range of commercial off-the-shelf (COTS) robots with a wide range of varying capabilities
(see fig.~\ref{fig:intro} (E)) that can be readily used for hierarchical
composition and real-world deployments. An appropriate choice of robot that considers several factors including the cost can be made using the metrics used in figure~\ref{fig:intro}. 

The type of behavioral architecture in figure~\ref{fig:intro} (A) will enable using collective behavior on simple robots without the need for accurate localization, as long as neighbor sensing is feasible. The higher-level robots could know mission goals with accurate localization, and the lower-level robots could be simpler robots performing collective behaviors using local sensing. The swarm composed of these types of robots can perform the mission in unison and has demonstrated that it can scale well with increasing mission complexity during our radiation cleanup task.

\subsubsection{Guide localization}
To achieve accurate positioning and obstacle detection on the guide robots in feature-poor environments, such as empty corridors, we leverage 2D Lidar Simultaneous Localization And Mapping (SLAM) from~\cite{hess20162Dlidar}, and we fuse it with visual odometry from an Intel Realsense T265 camera. The fusion of pose between Lidar odometry and visual odometry is done using a Kalman filter. The localization setup with a lidar and visual odometry has the advantages of both localization systems. Figure~\ref{fig:odom_diff} (A) shows the estimated odometry of the robot from visual odometry, and fused Kalmen filter estimates, (B) and (C) show the estimates in both coordinates over time. The localization accuracy of visual odometry is highly influenced by changes in illumination, scene, and features in the environment. In particular, when the robots navigate from a room into the corridors, there is a change in scene and illumination, causing the pose estimates to accumulate significant errors in the magnitude of meters. A similar error accumulates when the robot turns at the corridor's intersection—overall, resulting in errors ranging over 5 meters at the target. The visual SLAM system attempts to fix the estimate when it reaches back to the home location through map optimization; however, using vision-only localization might have more significant errors that endanger safe navigation in the experimental arena.

\begin{figure}
\centering
\includegraphics[width=0.5\textwidth]{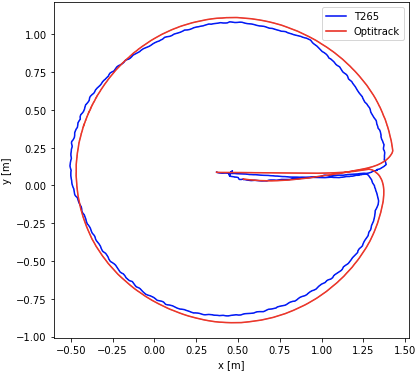}
\caption{Visual pose estimates (T265) are compared to motion capture system estimates (Optitrack) when the robot uses visual odometry to take a circular path around another robot inside a room with minimal scene change.}
\label{fig:odom_opti_t265} 
\end{figure}

\begin{figure}
\centering
\includegraphics[width=0.4\textwidth]{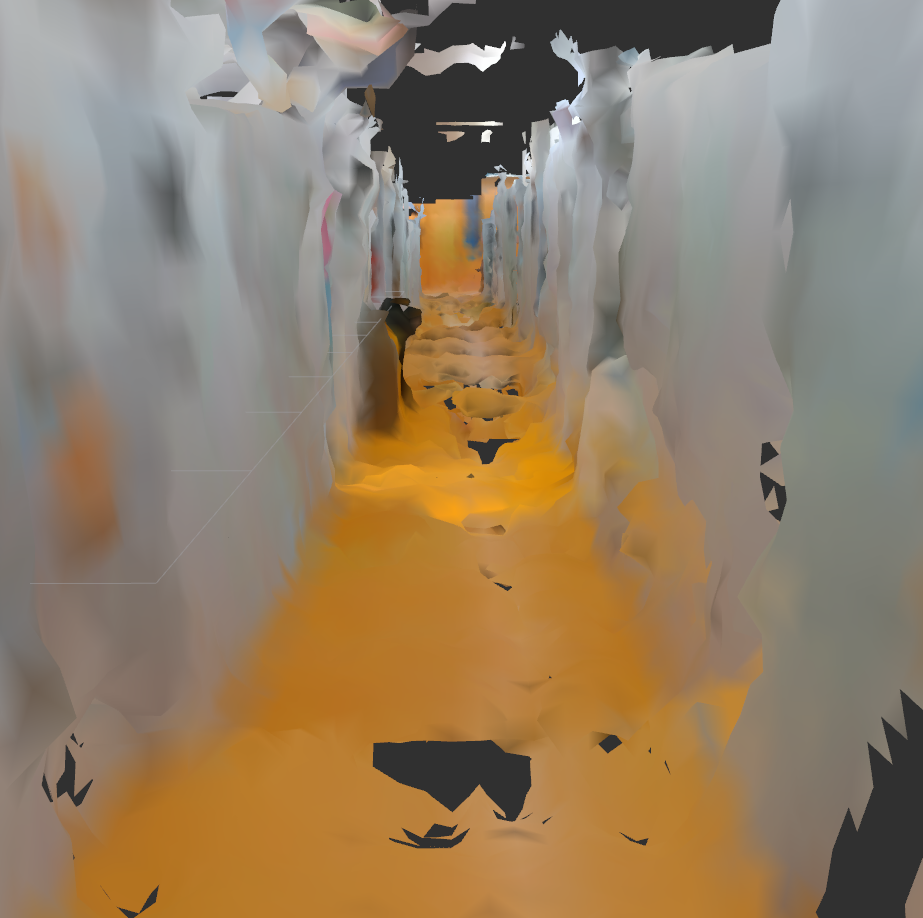}
\includegraphics[width=0.3\textwidth]{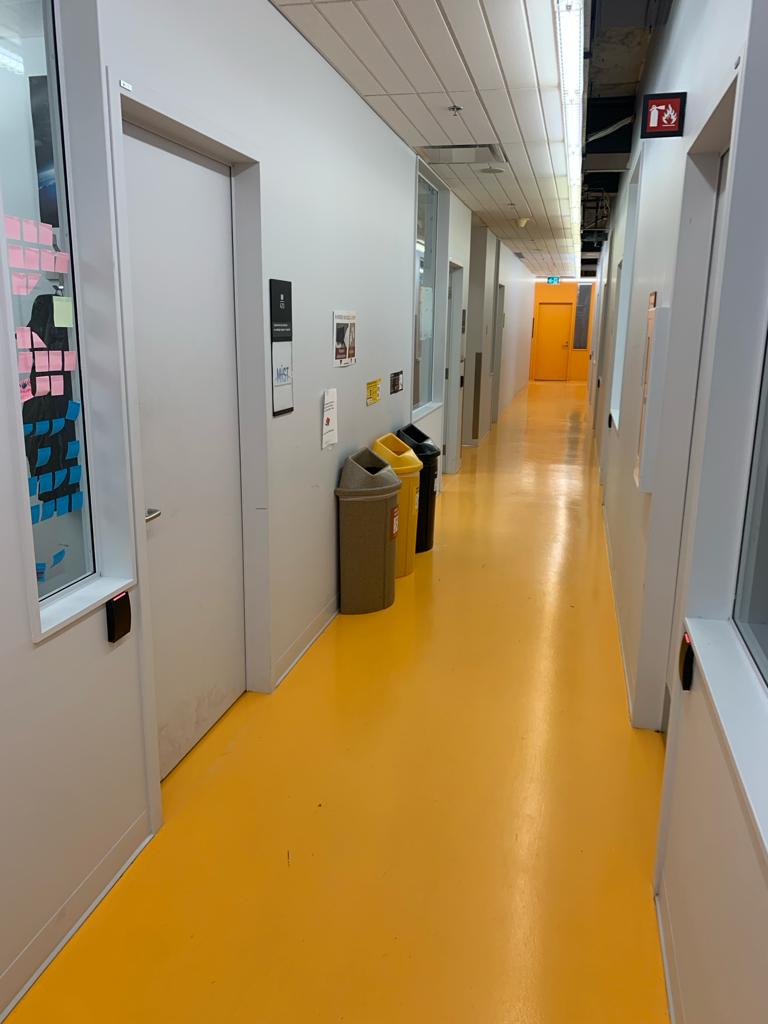}
\caption{The mesh constructed using the D435 camera on the robot and the image of the environment.}
\label{fig:mesh} 
\end{figure}

Figure~\ref{fig:odom_opti_t265} shows the localization estimates from two systems (visual and motion capture system) of a robot using visual odometry to navigate around another robot. During the navigation experiment, there were minimal scene and illumination changes as the robot stayed in a lab environment. Inside the lab environment, the errors were considerably minor in the range of 0.1m. With the errors induced through illumination and scene changes, achieving reliable performance with vision-only pose estimates might be difficult. The observed errors in the system led to the choice of a fused estimate from Lidar and vision for the guide robots. The fused pose estimate is used for mesh construction and motion control, as shown in Figure~\ref{fig:guidearc}. Figure~\ref{fig:mesh} shows the mesh constructed using the D435 stereo color camera compared to the actual environment during one of the experimental runs.

\subsubsection{Robot coordinate system}
\begin{figure}
\centering
\includegraphics[width=0.9\textwidth]{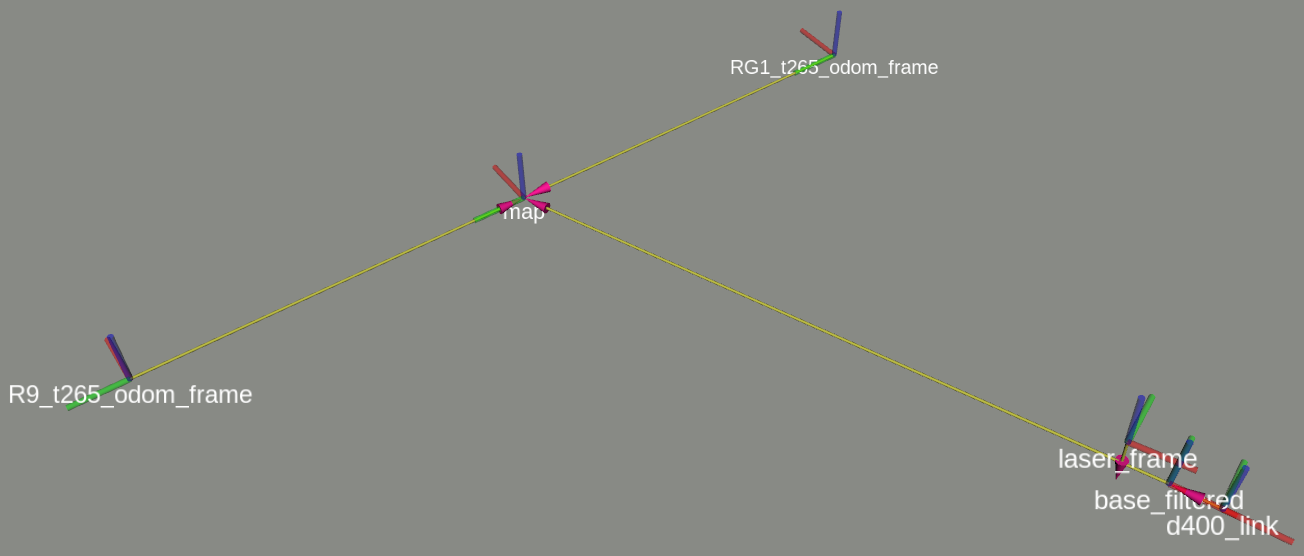}
\caption{A subset of the coordinate systems and its transformations on the guide robots used during the experimental evaluations.}
\label{fig:tf} 
\end{figure}

Figure~\ref{fig:tf} shows a subset of the coordinate systems and its transformation. At initialization, the guide robots calibrate the offset in the map of the robot and other guide robots from a configuration file. The coordinates indicate the transform between the map and other guides, named \textit{RG1\_t265\_odom\_frame} and \textit{R9\_t265\_odom\_frame} in figure~\ref{fig:tf}. In order to keep the offset configuration constant, the robots were started from a fixed deployment point. However, with techniques like TEASER~\cite{teaser}, scan matching could be made to determine the offset between the maps of the robots. The offset calibrated at initialization is used as a static transform to merge the map between robots. There was a set of other coordinates to represent the fixed frame of the map, robot base position, and sensors. The coordinate frames of the guide robots were designed based on the REP-105 coordinate frame standard in ROS for mobile platforms.
  
\subsubsection{Worker neighbor sensing}
The worker robots were installed with Intel Realsense T265 camera modules; the robots only used the left camera of the stereo pair. The Right camera images and pose estimates were not used. The worker robots used the camera images to sense the neighboring robots within the field of view. The workers directly measure the relative position of the neighbors with camera images. In order to facilitate the measurement of the relative position of neighboring robots, all the robots had an AprilTags~\cite{olson2011tags} behind the robot. The robots had a dedicated custom ROS Node that ran the software to detect the tags within the fisheye camera images. The following steps were performed to detect and extract the relative position of neighbors given an image: 
\begin{enumerate}
\item Undistort and Dewarp: The intrinsic and extensions of the camera are used to apply the \textit{initUndistortRectifyMap} within the opencv fisheye camera model, creating a map for camera images to be undistorted.
\item Resize image: The undistorted images are cropped to remove the edges in the fisheye image.
\item Run AprilTag Detector: The April tag detector is run on the resized images to obtain the list of detected tags and their pose. 
\item Transform the detection: Receive the list of detections and transform it to the mobile base coordinate frame from the camera frame.
\item Publish detection: The detections are published in ROS to be used by the control node.
\end{enumerate}

The steps described above are performed in a loop at 10Hz to obtain the relative position of the neighbors. AprilTags are known to provide high localization accuracy while being robust to lighting conditions, occlusions, distortions, warping, and view angles. The control node uses the detection published to localize itself relative to all the robots in the field of view and perform the required collective behaviors.

\begin{figure}
\centering
\includegraphics[width=0.7\textwidth]{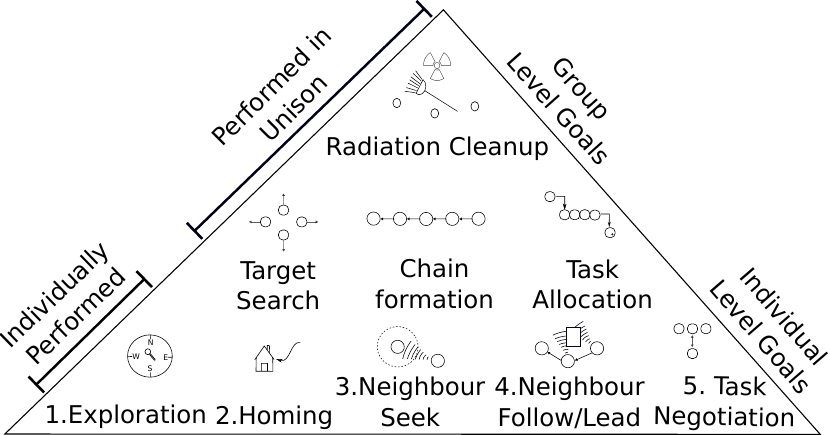}
\caption{Behavioral Hierarchy used during the radiation cleanup mission. The high-level behavior required by the mission is realized through a set of rudimentary behaviors performed by the robots on the lower levels.}
\label{fig:behaviorhierarchy} 
\end{figure}

\subsection{Coordination components}
Figure~\ref{fig:behaviorhierarchy} shows the behavioral hierarchy used to perform the radiation cleanup mission. As detailed in materials and methods, the behavioral hierarchy contains the radiation cleanup as the high-level swarm behavior that triggers a sequence of robot behaviors based on the mission progress. The mission progress dictates the robot's three coordinated behaviors: Target search, Chain formation, and Task allocation. The three coordinated behaviors require the robots to perform an individual local behavior coordinated among the robots. One of the main factors in coordinating information among the robots is obtaining the most updated information from all the robots through communication. 
 
The coordination of information in our system happens through gossip; the robots broadcast messages in the local neighborhood that are used to update local copies, and the updated information is rebroadcast in the network to be updated by other robots. The implementation of Virtual stigmergy~\cite{pinciroli2016tuple} in Buzz enables the realization of shared tuple memory between the robots, and it is gossip-based. The behaviors use Virtual stigmergy to share and update information used for coordination. 

\subsubsection{Target Search}
During target search, the robots perform exploration and homing based on the information coordinated with the swarm. The guide robots perform systematic frontier exploration, and the worker robots use SGBA~\cite{McGuire2019} to explore. Algorithm~\ref{algo:exp_home} shows the abstract sequence of actions performed by the robots during target search.
\begin{algorithm}
 \scriptsize
   \caption{Target Search behavior to identify target $t_i \in \mathbb{T}$}\label{algo:exp_home}
   \begin{algorithmic}[1]
     \State{List\_of\_targets=stigmergy.create() \# List of known targets shared among robots}
     \Procedure{Target Search}{}
     \label{algo:ts}
       \State{Update\_neighbor\_cmd()}
       \State{Update\_radiation\_distance()}
        \If {GuideRobot}
            \If { Neighbor\_cmd $\not=$ New target found by guide and Radiation\_distance $>$ Sensing distance and not Home}
                \If {Initialization not done}
                    \State{Set\_preferred\_exploration\_direction()}                    
                    \State{Perform\_initialization\_motion()}
                \Else
                    \If {Planner Not Triggered}                    
                        \State{Trigger\_Exploration\_planner()}
                    \Else
                        \State{Obtain\_path()}               
                        \State{Execute\_path()}
                    \EndIf
                 \EndIf       
             \ElsIf {Home}
                     \State {update\_neighbor\_cmd\_and\_broadcast()}
                     \If {Home Planner Not Triggered}                    
                        \State{Trigger\_Homing\_planner()}
                    \Else
                        \State{Obtain\_path()}               
                        \State{Execute\_path()}
                    \EndIf
              \ElsIf {Radiation\_distance $\leq$ Sensing\_distance}
                    \If {Target Unknown}                     
                        \State{List\_of\_targets.put(Target\_id,Target\_Location)}
                        \State{Home = True}    
                                      
                    \EndIf
              \ElsIf {Neighbor\_cmd == New target found by guide}
                    \State{Home = True}
              \EndIf
     
         \ElsIf {WorkerRobot}
            \If {Egalitarian or Heterogeneous Control}
             \If {Initialization not done}
                 \State {Set\_preferred\_exploration\_direction()}
             \ElsIf {Radiation\_distance $\leq$ Sensing\_distance}
                 \State {Target\_Seek()}
             \ElsIf {Neighbor Beckon in range}
                 \State {Neighbor\_Seek()}
             \ElsIf {Neighbor\_cmd == New target found by guide}
                 \State {SGBA\_Step(Inbound)} \# Homing
                 \If {Home reached}
                     \State {Update\_available\_for\_mobilization()}
                     \State {Switch\_state\_to\_chain\_formation()}
                 \EndIf
             \Else
                 \State {SGBA\_Step(Outbound)} \# Exploration
             \EndIf
            \Else 
                \State {\# Hierarchical Control}
                \State {Update\_available\_for\_mobilization()}
                \State {Switch\_state\_to\_chain\_formation()}
            \EndIf                 
         \EndIf
            
     \EndProcedure 
   \end{algorithmic}
 \end{algorithm}
 
The coordination between the robots is achieved through local broadcast command and a shared list of known targets with Virtual Stigmergy. The broadcast command include the neighbor command indicating the status of the robots.
Guide robots use the neighbor command to indicate the identification of a new target. The guide robots receiving the new target identification from another guide return home to determine the guide robots that will mobilize the workers to the target.
The worker receiving the guide neighbor command returns home for mobilization to target.
Upon identifying a new target through sensing, the guide robot updates the list of targets with the new target. The target information are initially set to be incomplete until the worker robots are mobilized to the target.
Once the robots arrive at the home location, all the guide robots continuously broadcast information to update the list of known targets and its status.

\subsubsection{Task allocation and Chain formation}
The task allocation mechanism on the robots enables determining the guide robots that will mobilize the workers to the target. The guide robots reach the home location, wait for a predefined period, and initiate task allocation. During task allocation, two sets of guide robots exist in the swarm: the guides that found a new target and the guides that did not find a new target. The guides that found the target take the chain leader role for chain formation and a guide from the other set without target detection is elected as a redundant guide to follow the chain. The robots determine the number of targets found by the guides during a task allocation and perform multiple rounds of task allocation by going through the list of all new target detections. The chain formation is initiated once the guide robots leading and following the chain are determined. In chain formation, the robots sequentially elect another robot to follow the electing robot until the required number of robots are in the chain. Once the required number of robots is determined, the robots perform chain movement to the target by following the guide with the target information.     

%\subsection{Adaptive Exploration Quota for Heterogeneous }
%The Heterogeneous approach is very similar to the Hierarchical strategy with the main difference of allowing the worker robots to explore. The 

%\begin{figure}
%  \centering
%  \includegraphics[scale=0.7]{figures/Plots/Bug_Exploration_Quota_Median.pdf}
%  \caption{Initial Exploration Quota for the Workers}
%  \label{fig:InitExplorationQuota}
%\end{figure}

\subsection{Communication between robots}
\label{sec:comms}
%\begin{figure}
%  \centering
%  \includegraphics[scale=0.7]{figures/Plots/delay_bet_robots_and_distance.pdf}
%  \caption{DelayVsDistance in the network}
%  \label{fig:delaygraph}
%\end{figure}
The communication in the simulation was realized using the Range and Bearing sensor that simulate situated communication in ARGoS3. In situated communication, a message's receiver can measure the sender's range and bearing. Range and bearing sensors are usually made of IR transceivers to send information while measuring the distance and angle. IR-based systems are prone to provide poor bearing estimation, and the message payloads are limited to a few bytes. The hardware deployment used in this study had a 5Ghz radio for communication and neighbor tag detection using the fisheye cameras to measure the relative position of neighbors.

\begin{figure}
\centering
\includegraphics[width=0.9\textwidth]{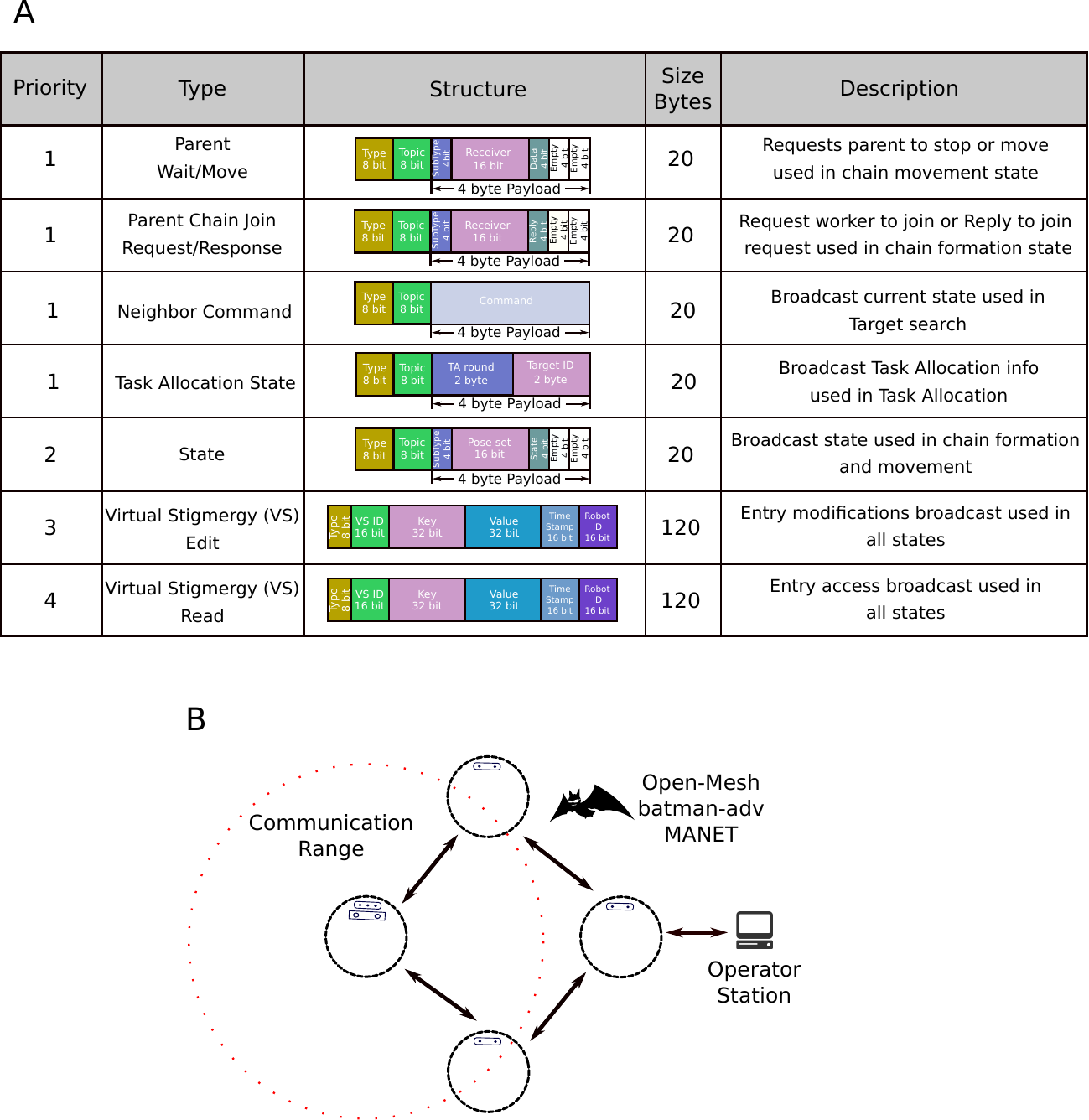}
\caption{(A) Types of communication messages and its structure used in the radiation cleanup mission. (B) Illustration of the communication setup used on the robots with Batman-adv.}
\label{fig:comms} 
\end{figure}

The communication on the hardware was a mesh network created using Independent Basic Service Set (IBSS) mode and a layer 2 routing protocol with Batman-adv. Figure~\ref{fig:comms} (B) illustrates an ad-hoc network with robots communicating peer-to-peer, and an operator connects to one of the robots to obtain the status information. We use the WiFi hardware that most off-the-shelf computers offer to create a MANET and have robots communicate peer-to-peer without requiring a router or additional radio. The ad hoc network is created using BATMAN-adv, providing a neighbor discovery. The robots used custom software that broadcast the serialized messages to other robots using UDP. The serialized messages contain the prioritized messages to be sent to neighboring robots to realize gossip-based communication. The message serialization and prioritization were performed within the Buzz Virtual Machine~\cite{pinciroli2016buzz} used to execute the control scripts. A message size limit of 500 bytes was set on the robots, both in simulation and hardware. In simulation, the communication range of robots was set to 10m, and there were no enforced limits on the hardware. However, the robots' hardware communication range was around 15m. The robots might still have connectivity beyond the range at reduced throughput.

Figure~\ref{fig:comms} (A) shows the robots' messages to coordinate with the swarm during the mission. The Neighbor command used during target search allows robots to signal each other of new target detection within the communication range. Task allocation state messages enable the robots to identify the current task allocation round and the target considered by the current task allocation round on each robot. The task allocation state messages are used to coordinate multiple rounds of task allocation on the robots. The parent wait/move messages are used during chain movement to request a robot to hold or move from the following robot. Parent Chain Join request and response are used during the chain formation to elect the robot that will follow a given robot. The status messages in chain formation are used to determine whether the robots are not committed to a chain and can be elected to join a chain. The messages indicating a payload of 4 bytes are classified as neighbor broadcasts as the message uses a common topic to broadcast and listen. The virtual stigmergy write and read messages enable the shared tuple-space memory between the robots. We refer the reader to~\cite{pinciroli2016tuple} for further details on virtual stigmergy. The virtual stigmergy tables are used for storing and sharing the following information: 
\begin{itemize}
\item List of Targets: The list of targets found by the guide robots.
\item Available guides: The list of available guides for task allocation.
\item Available workers: List of worker robots waiting to be mobilized to the target.
\item Cost negotiation: The table is used to determine the robot with the lowest cost to perform the task.
\end{itemize}
The task allocation is performed using the algorithm presented in~\cite{Varadharajan2020}, where the robot with the lowest cost is determined to perform a specific task. The cost negotiation table is used in task allocation to determine the robot with the lowest cost.

\section{Data analysis and statistics}

\subsection{Performance Metrics}
The performance metrics used to evaluate the performance of various control strategies are discussed in this section. 
\subsubsection{Mission success}
We assign costs to the robots to enable performance comparison across various control strategies with different types of robots. The worker robots are assigned a cost of one, and the guide robots are assigned a cost of three, using the operation cost determined through power consumption. We declare a mission to be successful when ten worker robots reach the target location to perform the cleanup operation. In Hierarchical control, the uniform cost swarm was composed of ten workers and two guides, and all the ten worker robots were required to reach the target location to complete the mission. With Egalitarian control, there were sixteen worker robots, and ten workers had to reach the target to succeed. Heterogeneous control had ten workers and two guides; all ten workers were required to reach the target location. The percentage of successful missions indicates the number of missions with ten worker robots at the target location. The urban environment had 90 experimental runs, the maze had 30 experimental runs, and the forest had 30 experimental runs.    

\subsubsection{Normalized time}
The normalized time is the metric used to evaluate the time a given control strategy takes to complete the mission. The time taken during the experimental evaluations is normalized with the most significant mission time of Hierarchical control. The normalization time of the Hierarchical control, determined to be 3.33 hours, is set as the experimental time limit for the other control strategies. The normalized time metric was only computed for the successful experiments, and the failed missions did not have a reported time.

\subsubsection{Percentage of explored area}
The percentage of explored areas indicates the proportion of the environment explored by the robots before completing the mission. The percentage of the area explored was computed using the following procedure:
\begin{itemize}
\item Creating grid map: The map of the environment indicating the obstacle locations is used to create a discretized grid map. 
\item Update visitation: Robot positions obtained during the experiments are used to update the discretized grid to indicate if a particular grid cell is visited.
\item Compute coverage: The ratio between the number of visited and free cells provides the explored area. 
\end{itemize}
Computing the percentage of explored areas was performed using the python package \textit{pandas}.

\subsubsection{Average neighbor distance and count}
The average neighbor distance measure is used in the Egalitarian control strategy to identify the average distance from a given robot to its neighbors. We define a robot to be a neighbor if the robot is within the communication range of any given robot. The average neighbor distance is computed using the following steps:
\begin{enumerate}
\item compute the distance of all the neighbors for a given robot during each experimental step.
\item combine each robot's neighbor distances list to create a single list of neighbor distances.
\item compute the average neighbor distance for an experiment by averaging the list of neighbor distances.
\end{enumerate}
The average neighbor count measure is computed similarly to the average neighbor distance. A combined list of neighbors is created from each robot neighbor count list and averaged to obtain the average neighbor count.    

\subsubsection{Incremental coverage}
Incremental coverage measure shows the amount of new environment the robots cover at a given experimental time. The incremental coverage was computed using a similar procedure for computing the explored area percentage. Unlike the percentage of explored area, the coverage is obtained for a specific time window in incremental coverage. 

\subsubsection{Average time to find targets}
The average time to find the target metric is used with Hierarchical control experiments with multiple targets. It indicates the robot's average time to identify the targets in the environment when multiple targets are placed symmetrically in the environment. The metric is computed by finding an average of all the available target detection times from all the robots.

\subsection{Statistics}

Our primary data organization and analysis tool was the \texttt{pandas} Python package; all the statistics derived were using this package. Table~\ref{tab:hir_time_data} shows the data statistics of the normalized time for Hierarchical control. Table~\ref{tab:hir_coverage_data} and \ref{tab:bug_coverage_data} show the percentage of explored area statistics for Hierarchical and Egalitarian control.

We performed statistical tests to understand the significance of results obtained (table~\ref{tab:hir_time_data}-\ref{tab:bug_stat_model}). The statistical analysis was performed using the Python package \texttt{statsmodels}. At first, ANOVA statistical tests were performed on the time and percentage of explored area metrics using a statistical model derived from the data. Upon obtaining a significance on the data (small p-values), post-hoc testing was performed using Tukey HSD. The results reveal that both the time taken and explored area had significant differences in data distributions for various arena sizes, map types (except for indoor and urban), and different guide numbers (except for 6-10 guides). These results highlight the fact that the urban environment had some overlap with that of the maze environments. Furthermore, the trends displayed by 6-10 guide robots had overlaps and can be justified considering the fact that additional robots involved in the exploration task does not improve performance in the same proportion. The diminishing marginal utility can again be verified with the statistical results. 
\begin{table}[htbp]
  \caption{Data statistics of normalized time to complete the mission when using Hierarchical control across various experimental configurations.}\label{tab:hir_time_data}
\small
\centering
\begin{adjustbox}{max width=\textwidth}
 \begin{tabular}{|c|c|c|c|c|c|c|c|c|c|c|} 
 \hline
% Header 
    Arena type & Arena Size & Guides Num. & Count & Mean & std & min & 25\% & 50\% & 75\% & max \\
 %  \multirow{2}{*}{Map type} & \multicolumn{5}{c|}{Size (Bytes)} \\
  % \cline{2-6} 
 %  &   ID & Key & Value & Extra & Total \\ 
 \hline
% Urban Entry 
\multirow{15}{*}{Urban}  & \multirow{5}{*}{30} & 2 &   90.0 &  0.043827 &  0.049934 &  0.011515 &  0.027448 &  0.035570 &  0.046286 &  0.475899 \\
  &  & 4 &   90.0 &  0.023649 &  0.015429 &  0.010316 &  0.013160 &  0.021998 &  0.028243 &  0.126635 \\
  &  & 6 &   90.0 &  0.017747 &  0.007513 &  0.010033 &  0.011948 &  0.015254 &  0.021465 &  0.042439 \\
  &  & 8 &   90.0 &  0.016460 &  0.007032 &  0.010116 &  0.011905 &  0.013863 &  0.017117 &  0.042431 \\
  &  & 10&	 90.0 &  0.016013 &  0.005413 &  0.009958 &  0.012117 &  0.013934 &  0.018447 &  0.037618 \\
  \cline{2-11} 
 &\multirow{5}{*}{60}& 2 &   90.0 &  0.110281 &  0.059502 &  0.026228 &  0.070678 &  0.097872 &  0.142364 &  0.336475 \\
  &  & 4 &   90.0 &  0.062298 &  0.030821 &  0.026178 &  0.037987 &  0.051836 &  0.081167 &  0.181322 \\
  &  & 6 &   90.0 &  0.050200 &  0.022467 &  0.023638 &  0.031961 &  0.047414 &  0.063091 &  0.128958 \\
  &  & 8 &   90.0 &  0.037483 &  0.015436 &  0.023089 &  0.028474 &  0.031457 &  0.040289 &  0.102089 \\
  &  & 10&	90&	0.038625 &  0.017726 &  0.024338 &  0.027529 &  0.030578 &  0.040797 &  0.108875 \\
 \cline{2-11}
 &\multirow{5}{*}{120} & 2 &   90.0 &  0.294785 &  0.203918 &  0.063497 &  0.173050 &  0.212788 &  0.340218 &  1.000000 \\
  &  & 4 &   90.0 &  0.181650 &  0.117935 &  0.062023 &  0.098990 &  0.149870 &  0.224796 &  0.618671 \\
  &  & 6 &   90.0 &  0.136383 &  0.065771 &  0.051731 &  0.086146 &  0.126398 &  0.164108 &  0.390080 \\
  &  & 8 &   90.0 &  0.103684 &  0.050056 &  0.049683 &  0.064229 &  0.091215 &  0.128048 &  0.321213 \\
  &  & 10&	90&	0.096148 &  0.052164 &  0.051315 &  0.064641 &  0.080828 &  0.113950 &  0.431420 \\
 \hline
% indoor entry
 \multirow{15}{*}{Indoor}  & \multirow{5}{*}{30} & 2 &   30.0 &  0.052985 &  0.016128 &  0.013780 &  0.044727 &  0.055778 &  0.063719 &  0.085353 \\
  &  & 4 &   30.0 &  0.031785 &  0.014217 &  0.011574 &  0.017481 &  0.032518 &  0.042323 &  0.064804 \\
  &  & 6 &   30.0 &  0.028509 &  0.017406 &  0.011665 &  0.015997 &  0.023534 &  0.029846 &  0.084471 \\
  &  & 8 &   30.0 &  0.026534 &  0.011924 &  0.011557 &  0.017031 &  0.023588 &  0.036819 &  0.057718 \\
  &  & 10&	30&	0.025286 &  0.010628 &  0.011499 &  0.015221 &  0.025325 &  0.031634 &  0.050050 \\
 \cline{2-11}
& \multirow{5}{*}{60} &	2 &   30.0 &  0.116636 &  0.036692 &  0.057776 &  0.095674 &  0.108588 &  0.130865 &  0.245069 \\
  &  & 4 &   30.0 &  0.077310 &  0.025970 &  0.035687 &  0.061171 &  0.074775 &  0.092470 &  0.145145 \\
  &  & 6 &   30.0 &  0.055010 &  0.016145 &  0.032956 &  0.040624 &  0.054009 &  0.066026 &  0.097926 \\
  &  & 8 &   30.0 &  0.045917 &  0.017231 &  0.029833 &  0.033778 &  0.040204 &  0.058157 &  0.096119 \\
  & & 10&	30&	0.044128 &  0.015043 &  0.029925 &  0.031632 &  0.036503 &  0.052187 &  0.083305 \\
 \cline{2-11}
& \multirow{5}{*}{120} &	2 &   30.0 &  0.342714 &  0.150238 &  0.102347 &  0.257600 &  0.302754 &  0.423987 &  0.782250 \\
  &  & 4 &   30.0 &  0.246209 &  0.139942 &  0.066469 &  0.135840 &  0.215714 &  0.300141 &  0.622493 \\
  &  & 6 &   30.0 &  0.204239 &  0.107932 &  0.067235 &  0.127553 &  0.189874 &  0.238545 &  0.495175 \\
  &  & 8 &   30.0 &  0.165688 &  0.078720 &  0.068642 &  0.117416 &  0.144928 &  0.188639 &  0.365476 \\
  & & 10&	30&	0.111570 &  0.054916 &  0.064338 &  0.074285 &  0.093351 &  0.132790 &  0.306326 \\
 \hline
% Forest entry 
 \multirow{15}{*}{Forest}  & \multirow{5}{*}{30}& 2 &   30.0 &  0.033828 &  0.015519 &  0.014171 &  0.023143 &  0.033043 &  0.037702 &  0.093704 \\
  &  & 4 &   30.0 &  0.018382 &  0.008391 &  0.010966 &  0.012140 &  0.015791 &  0.019309 &  0.043289 \\
  &  & 6 &   30.0 &  0.016486 &  0.008909 &  0.010092 &  0.011759 &  0.012897 &  0.015499 &  0.042739 \\
  &  & 8 &   30.0 &  0.014534 &  0.004980 &  0.010383 &  0.011832 &  0.012485 &  0.013595 &  0.030782 \\
  & & 10&	30&	0.013635 &  0.003179 &  0.010183 &  0.011742 &  0.012623 &  0.015027 &  0.023888 \\
 \cline{2-11}
& \multirow{5}{*}{60}& 2 &   30.0 &  0.133836 &  0.072139 &  0.031648 &  0.094464 &  0.113047 &  0.148129 &  0.348765 \\
  &  & 4 &   30.0 &  0.086843 &  0.052065 &  0.028043 &  0.052962 &  0.080020 &  0.104196 &  0.289265 \\
  &  & 6 &   30.0 &  0.045730 &  0.023561 &  0.025753 &  0.028272 &  0.038397 &  0.056971 &  0.136901 \\
  &  & 8 &   30.0 &  0.050312 &  0.028652 &  0.024646 &  0.030235 &  0.039875 &  0.058001 &  0.122530 \\
  & & 10&	30&	0.042638 &  0.020353 &  0.023971 &  0.026994 &  0.034883 &  0.049021 &  0.101848 \\
 \cline{2-11}
& \multirow{5}{*}{120}&	 2 &   30.0 &  0.534111 &  0.258129 &  0.088542 &  0.315149 &  0.561477 &  0.675958 &  1.000000 \\
  &  & 4 &   30.0 &  0.310398 &  0.205737 &  0.069808 &  0.152438 &  0.274040 &  0.398906 &  1.000000 \\
  &  & 6 &   30.0 &  0.254202 &  0.212404 &  0.064396 &  0.121269 &  0.209794 &  0.269956 &  0.963289 \\
  &  & 8 &   30.0 &  0.159634 &  0.086941 &  0.055711 &  0.091510 &  0.136106 &  0.201978 &  0.369381 \\
  & &  10&	30  &	 0.114737 &  0.054339 &  0.054554 &  0.068732 &  0.096336 &  0.131670 &  0.240631 \\
\hline
 \end{tabular}
\end{adjustbox}
\end{table}
\begin{table}[htbp]
  \caption{Data statistics of explored area when using hierarchical control across various experimental configurations.}\label{tab:hir_coverage_data}
\small
\centering
\begin{adjustbox}{max width=\textwidth}
 \begin{tabular}{|c|c|c|c|c|c|c|c|c|c|c|} 
 \hline
% Header 
    Arena type & Arena Size & Guides Num. & Count & Mean & std & min & 25\% & 50\% & 75\% & max \\
 \hline
% Urban Entry 
\multirow{15}{*}{Urban}  & \multirow{5}{*}{30}&	10&	90&	0.899105&	0.070983&	0.686747&	0.861553&	0.916402&	0.954246&	0.993939\\
&&2&	90&	0.735332&	0.172519&	0.285714&	0.623624&	0.740200&	0.874651&	1.000000\\
&&4&	90&	0.821980&	0.123898&	0.503900&	0.721079&	0.837187&	0.925665&	1.000000\\
&&6&	90&	0.876241&	0.093334&	0.496825&	0.836040&	0.904761&	0.937192&	1.000000\\
&&8&	90&	0.891497&	0.069542&	0.661972&	0.832695&	0.902400&	0.940530&	1.000000\\
\cline{2-11}
& \multirow{5}{*}{60}&	10&	90&	0.849532&	0.084591&	0.676677&	0.788849&	0.840297&	0.914529&	1.000000\\
&&2&	90&	0.626288&	0.219874&	0.192246&	0.443852&	0.601401&	0.804303&	0.993451\\
&&4&	90&	0.724191&	0.148268&	0.457735&	0.591793&	0.704820&	0.845983&	0.992279\\
&&6&	90&	0.802260&	0.127362&	0.446688&	0.704434&	0.819407&	0.899634&	0.999618\\
&&8&	90&	0.819047&	0.095569&	0.617978&	0.765020&	0.820573&	0.890500&	0.998440\\
\cline{2-11}
& \multirow{5}{*}{120}&	10&	90&	0.742686&	0.099868&	0.556552&	0.674747&	0.722446&	0.791685&	0.996968\\
&&2&	90&	0.513280&	0.226678&	0.231000&	0.321454&	0.438710&	0.651586&	0.986546\\
&&4&	90&	0.608529&	0.180389&	0.347167&	0.435318&	0.586064&	0.745426&	0.979230\\
&&6&	90&	0.674769&	0.150436&	0.327153&	0.557075&	0.671861&	0.774278&	0.983364\\
&&8&	90&	0.705560&	0.123439&	0.435506&	0.621839&	0.693291&	0.804856&	0.971374\\
\hline
% indoor stats
\multirow{15}{*}{Indoor}  & \multirow{5}{*}{30}&	10&	30&	0.780826&	0.158541&	0.429027&	0.651960&	0.826224&	0.923135&	0.980892\\
&&2&	30&	0.749301&	0.150887&	0.405104&	0.652889&	0.800441&	0.863888&	0.982229\\
&&4&	30&	0.754557&	0.184854&	0.400319&	0.621886&	0.806987&	0.931796&	0.987055\\
&&6&	30&	0.818138&	0.154658&	0.408293&	0.733399&	0.877759&	0.942623&	0.972492\\
&&8&	30&	0.815510&	0.155555&	0.432217&	0.743123&	0.846201&	0.937482&	0.998382\\
\cline{2-11}
& \multirow{5}{*}{60}&	10&	30&	0.826724&	0.070785&	0.664644&	0.785567&	0.834541&	0.869657&	0.977346\\
&&2&	30&	0.619179&	0.110870&	0.394417&	0.522176&	0.649579&	0.690117&	0.900883\\
&&4&	30&	0.752980&	0.068117&	0.619337&	0.710839&	0.746807&	0.796475&	0.891181\\
&&6&	30&	0.788951&	0.048988&	0.699919&	0.744316&	0.788569&	0.827696&	0.876202\\
&&8&	30&	0.818410&	0.074460&	0.562195&	0.788661&	0.836510&	0.858751&	0.951535\\
\cline{2-11}
& \multirow{5}{*}{120}&	10&	30&	0.726601&	0.088633&	0.582735&	0.656098&	0.714862&	0.809160&	0.913228\\
&&2&	30&	0.533572&	0.113187&	0.282565&	0.464974&	0.532726&	0.609613&	0.752427\\
&&4&	30&	0.634216&	0.134227&	0.380048&	0.538975&	0.635338&	0.759085&	0.888933\\
&&6&	30&	0.718520&	0.106172&	0.534009&	0.651248&	0.692636&	0.795254&	0.918840\\
&&8&	30&	0.727304&	0.098267&	0.550687&	0.654849&	0.748194&	0.802894&	0.890451\\
\hline
% Forest stats
\multirow{15}{*}{Forest}  & \multirow{5}{*}{30}&	10&	30&	0.911512&	0.059290&	0.800380&	0.864017&	0.927003&	0.957823&	0.994186\\
&&2&	30&	0.772309&	0.183895&	0.459662&	0.558363&	0.841125&	0.905920&	0.996124\\
&&4&	30&	0.799273&	0.116915&	0.633397&	0.707373&	0.771100&	0.906947&	0.998073\\
&&6&	30&	0.875061&	0.087362&	0.615970&	0.844084&	0.883828&	0.925880&	0.994253\\
&&8&	30&	0.891957&	0.075676&	0.705441&	0.847459&	0.903664&	0.949860&	0.998062\\
\cline{2-11}
& \multirow{5}{*}{60}&	10&	30&	0.909783&	0.063164&	0.766160&	0.859185&	0.905814&	0.968768&	0.999034\\
&&2&	30&	0.722664&	0.219449&	0.223823&	0.609179&	0.742512&	0.920008&	0.997583\\
&&4&	30&	0.824659&	0.161100&	0.433237&	0.739527&	0.879989&	0.946346&	0.999041\\
&&6&	30&	0.784040&	0.133343&	0.533820&	0.672227&	0.795782&	0.895731&	0.993305\\
&&8&	30&	0.882209&	0.108604&	0.628710&	0.800456&	0.927190&	0.966968&	0.999034\\
\cline{2-11}
& \multirow{5}{*}{120}&	10&	30&	0.787428&	0.112893&	0.561366&	0.709728&	0.790648&	0.877123&	0.965798\\
&&2&	30&	0.664517&	0.235158&	0.146142&	0.461677&	0.758812&	0.824138&	0.987143\\
&&4&	30&	0.725772&	0.172165&	0.353118&	0.578523&	0.745085&	0.884574&	0.987360\\
&&6&	30&	0.774987&	0.163252&	0.478287&	0.659606&	0.819945&	0.906747&	0.999880\\
&&8&	30&	0.808961&	0.126433&	0.532179&	0.701991&	0.838459&	0.913940&	0.997113\\
\hline
 \end{tabular}
\end{adjustbox}
\end{table}
\begin{table}[htbp]
  \caption{Data statistics on explored area with egalitarian control across various experimental configurations.}\label{tab:bug_coverage_data}
\small
\centering
  \begin{adjustbox}{max width=\textwidth}
 \begin{tabular}{|c|c|c|c|c|c|c|c|c|c|} 
 \hline
% Header 
    Arena type & Arena Size & Count & Mean & std & min & 25\% & 50\% & 75\% & max \\
 \hline
% Urban Entry 
\multirow{4}{*}{Urban}  &15.0     &   90.0 &  0.999495 &  0.004087 &  0.961832 &  1.000000 &  1.000000 &  1.000000 &  1.000000 \\
 & 30.0     &   90.0 &  0.991398 &  0.015427 &  0.905901 &  0.990610 &  0.998423 &  1.000000 &  1.000000 \\
 & 60.0     &   90.0 &  0.928395 &  0.052696 &  0.663974 &  0.911287 &  0.941308 &  0.957125 &  0.993254 \\
 & 120.0    &   90.0 &  0.804201 &  0.062793 &  0.639101 &  0.770688 &  0.804162 &  0.845266 &  0.924279 \\
\hline
% Indoor stats
\multirow{4}{*}{Indoor}  &15.0     &   30.0 &  0.998907 &  0.004683 &  0.975410 &  1.000000 &  1.000000 &  1.000000 &  1.000000 \\
 & 30.0     &   30.0 &  0.994875 &  0.011950 &  0.943820 &  0.994815 &  1.000000 &  1.000000 &  1.000000 \\
 & 60.0     &   30.0 &  0.896739 &  0.053773 &  0.741573 &  0.874354 &  0.900354 &  0.930105 &  0.985841 \\
 & 120.0    &   30.0 &  0.659795 &  0.089660 &  0.502404 &  0.594800 &  0.666511 &  0.723514 &  0.850829 \\
\hline
% Forest stats
\multirow{4}{*}{Forest}  &15.0     &   30.0 &  0.984254 &  0.038048 &  0.828358 &  0.992423 &  1.000000 &  1.000000 &  1.000000 \\
 & 30.0     &   30.0 &  0.969228 &  0.168176 &  0.078799 &  1.000000 &  1.000000 &  1.000000 &  1.000000 \\
 & 60.0     &   30.0 &  0.965253 &  0.024131 &  0.896700 &  0.959398 &  0.969033 &  0.983727 &  0.991354 \\
 & 120.0    &   30.0 &  0.667676 &  0.040170 &  0.587265 &  0.645065 &  0.662256 &  0.698900 &  0.750060 \\
\hline
 \end{tabular}
 \end{adjustbox}
\end{table}

\begin{table}[htbp]
  \caption{Statistical test results on hierarchical method simulation data. The study involves multiple criteria (independent variables) and statistical tests are performed on two performance metric (normalized time and explored area). We fit a model that describes the data and performed ANOVA test on this model. Table (A) reports the model parameters for time statistics. Table (B) reports the Tukey Honestly Significant Difference (HSD) post hoc testing results for a significance of 95\% ($\alpha=0.05$) on time statistics.}
  \label{tab:hir_stat_model_time}
\small
\centering
  \begin{adjustbox}{max width=\textwidth}
 \begin{tabular}{|c|c|c|c|c|} 
 \hline
 \multicolumn{5}{|c|}{(A) Normalized time ANOVA model parameters}\\
 \hline
% Header 
    &sum\_sq &	df&	F & P-Value($>$F) \\
 \hline
C(arena size)&	11.985281&	2&	758.172167&	3.280524e-252\\
C(arena type)&	0.601136&	2&	38.027009&	5.742943e-17\\
C(guide num.)&	3.959847&	4&	125.247209&	1.351431e-96\\
Residual&	17.713004&	2241&	-&	-\\
\hline
 \end{tabular}

 \vspace*{0.5cm}
  \begin{tabular}{|c|c|c|c|c|c|} 
 \hline
  \multicolumn{6}{|c|}{(B) Normalized time post-hoc testing}\\
 \hline
   \multicolumn{6}{|c|}{Arena size}\\
 \hline
% Header 
    arena size&	arena size&	mean diff&	lower&	upper&	reject\\
 \hline

120&	30&	-0.1707&	-0.1827&	-0.1586&	True\\
120&	60&	-0.1315&	-0.1435&	-0.1194&	True\\
30&	60&	0.0392&	0.0271&	0.0513&	True\\
\hline
   \multicolumn{6}{|c|}{Arena type}\\
 \hline
 arena type&	arena type&	mean diff&	lower&	upper&	reject\\
 \hline
forest&	indoor&	-0.017&	-0.0361&	0.0021&	False\\
forest&	urban&	-0.04&	-0.0556&	-0.0244&	True\\
indoor&	urban&	-0.023&	-0.0386&	-0.0074&	True\\
\hline
   \multicolumn{6}{|c|}{Number of Guides}\\
 \hline
 guide num. & guide num.&	mean diff&	lower&	upper&	reject\\
 \hline
10&	2&	0.1171&	0.096&	0.1382&	True\\
10&	4&	0.0513&	0.0301&	0.0724&	True\\
10&	6&	0.0275&	0.0064&	0.0487&	True\\
10&	8&	0.0087&	-0.0124&	0.0299&	False\\
2&	4&	-0.0658&	-0.0869&	-0.0447&	True\\
2&	6&	-0.0896&	-0.1107&	-0.0684&	True\\
2&	8&	-0.1084&	-0.1295&	-0.0872&	True\\
4&	6&	-0.0238&	-0.0449&	-0.0026&	True\\
4&	8&	-0.0425&	-0.0637&	-0.0214&	True\\
6&	8&	-0.0188&	-0.0399&	0.0024&	False\\
\hline
 \end{tabular}
\end{adjustbox}
\end{table}
\begin{table}[htbp]
  \caption{ ANOVA model parameters and post-hoc testing results for hierarchical control simulations. Table (A) reports the ANOVA model parameters for explored area. Table (B) reports the Tukey Honestly Significant Difference (HSD) post hoc testing results for a significance of 95\% ($\alpha=0.05$) on explored area.}\label{tab:hir_stat_model_explored_area}
\small
\centering
  \begin{adjustbox}{max width=\textwidth}
  \begin{tabular}{|c|c|c|c|c|} 
 \hline
 \multicolumn{5}{|c|}{(A) Explored area ANOVA statistics model parameters}\\
 \hline
% Header 
    &sum\_sq &	df&	F &P-Value($>$F) \\
 \hline
C(arena size)&	6.847954e-10&	2&	253.394023&	6.217476e-100\\
C(arena type)&	9.489673e-11&	2&	35.114525&	9.641286e-16\\
C(guide num.)&	6.898546e-10&	4&	127.633043&	2.824571e-98\\
Residual&	3.026791e-09&	2240&	-&	-\\
\hline
 \end{tabular}
  \vspace*{0.5cm}
  
  \begin{tabular}{|c|c|c|c|c|c|} 
 \hline
  \multicolumn{6}{|c|}{(B) Explored Area statistics post-hoc testing}\\
 \hline
   \multicolumn{6}{|c|}{Arena size}\\
 \hline
% Header 
    arena size&	arena size&	mean diff&	lower&	upper&	reject\\
 \hline
120&	30&	1.33409540e-06&	1.17625665e-06&	1.49193416e-06&	True\\
120&	60&	8.51880085e-07&	6.94041331e-07&	1.00971884e-06&	True\\
30&	60&	0.0392&-6.40001414e-07&	-3.24429220e-07&	True\\
\hline
   \multicolumn{6}{|c|}{Arena type}\\
 \hline
 arena type&	arena type&	mean diff&	lower&	upper&	reject\\
 \hline
forest&	indoor&	-5.97025110e-07&	-8.15987136e-07 & -3.78063084e-07&	True\\
forest&	urban&	 -4.71803063e-07&	-6.50634519e-07 & -2.92971608e-07&	True\\
indoor&	urban&	1.25222046e-07&	-5.34602380e-08 &  3.03904331e-07&	False\\
\hline
   \multicolumn{6}{|c|}{Number of Guides}\\
 \hline
 guide num. &	guide num. &	mean diff&	lower&	upper&	reject\\
 \hline
10&	2&	-1.51660406e-06&	-1.75379280e-06 & -1.27941532e-06&	True\\
10&	4&	-8.11426430e-07&	-1.04848321e-06 & -5.74369646e-07&	True\\
10&	6&	-3.31690903e-07&	-5.68747687e-07 & -9.46341188e-08&	True\\
10&	8&	-1.24483519e-07& -3.61540303e-07  & 1.12573265e-07&	False\\
2&	4&	 7.05177628e-07&4.67988889e-07  & 9.42366367e-07&	True\\
2&	6&	1.18491316e-06&	9.47724417e-07 &  1.42210189e-06&	True\\
2&	8&	1.39212054e-06&	1.15493180e-06  & 1.62930928e-06&	True\\
4&	6&	4.79735527e-07&	2.42678743e-07  & 7.16792311e-07&	True\\
4&	8&	6.86942911e-07&	4.49886127e-07 &  9.23999695e-07&	True\\
6&	8&	2.07207384e-07&	-2.98494003e-08  & 4.44264168e-07&	False\\
\hline
 \end{tabular}
 \end{adjustbox}
\end{table}

\begin{table}[htbp]
  \caption{Statistical test on the egalitarian control simulation data. The various experimental configurations creating independent variables is tested for significance using ANOVA tests. Table (A) reports the ANOVA model parameters for explored area. Table (B) post-hoc testing for a significance of 95\% ($\alpha=0.05$) on explored area.}\label{tab:bug_stat_model}
\small
\centering
  \begin{adjustbox}{max width=\textwidth}
  \begin{tabular}{|c|c|c|c|c|} 
 \hline
 \multicolumn{5}{|c|}{(A) Explored area ANOVA statistics model parameters}\\
 \hline
% Header 
    &sum\_sq &	df&	F &P-value($>$F) \\
 \hline
C(arena size)&	5.99786880&	3&	485.26587254&	2.82999030e-159\\
C(arena type)&	0.22145999&	2&	26.87629049&	6.69970962e-12\\
Residual&	2.44727291&	594&	-&	-\\
\hline
 \end{tabular}
  \vspace*{0.5cm}
  
  \begin{tabular}{|c|c|c|c|c|c|} 
 \hline
  \multicolumn{6}{|c|}{(B) Explored Area statistics post-hoc testing}\\
 \hline
   \multicolumn{6}{|c|}{Arena size}\\
 \hline
% Header 
    arena size &	arena Size &	mean diff&	lower&	upper&	reject\\
 \hline
15&	30&	-0.0087&	-0.0286&	0.0112&	False\\
15&	60&	-0.0669&	-0.0868&	-0.047&	True\\
15&	120&	-0.2483&	-0.2682&	-0.2284&	True\\
30&	60&	-0.0582&	-0.0781&	-0.0383&	True\\
30&	120&	-0.2396&	-0.2596&	-0.2197&	True\\
60&	120&	-0.1814&	-0.2013&	-0.1615&	True\\
\hline
   \multicolumn{6}{|c|}{Arena type}\\
 \hline
 arena type &	arena type &	mean diff&	lower&	upper&	reject\\
 \hline
Forest&	Maze&	-0.009&	-0.0451&	0.0271&	False\\
Forest&	Urban&	0.0343 	&	0.0048&	0.0637&	True\\
Maze&	urban&	0.0433&	0.0138 &	0.0727&	True\\
\hline
 \end{tabular}
 \end{adjustbox}
\end{table}

\newpage
\section*{Supplementary video captions}
\subsection*{Video 1: Heterogeneous control mission}
The video shows a Heterogeneous control strategy with ten workers and two guides in an urban environment. The worker robots explore using SGBA to identify the target, and the guide robots use frontier exploration to identify the target. Workers return to the home location after some time, and the guide robots return home upon identification of the target or when another guide finds the target. The guides at the home location mobilize the available worker robots to the target location by forming a chain.
\subsection*{Video 2: Hierarchical control mission}
The video shows the Hierarchical control strategy with ten workers and two guides in a maze environment. The guide robots explore the environment and identify the target. Upon identifying the target, the guides return home and mobilize the worker robots to the target location. 
\subsection*{Video 3: Egalitarian control mission}
The video shows the Egalitarian control strategy with 25 worker robots in an urban environment. The worker robots use local sensing to explore the environment and identify the target. The workers at the target location act as beacons to attract other workers to the target location. 

\subsection*{Video 4: Scalability experiment}
This video demonstrates the scalability experiment carried out with the Hierarchical control strategy. It shows the setting with 4 targets for an urban $ 30 m \times 30 m$ environment with six guides and 25 workers. The video's experimental configuration differed from the experiments used in the paper to demonstrate the flexibility of the simulation stack. The guide robots mobilized four worker robots instead of ten workers, and only one guide mobilized the workers instead of two guides.     
\subsection*{Video 5: Hardware experiments}
The summary of hardware experiments shows Hierarchical control with six worker robots and two guides. The guide robots explore the corridor environment and identify the target. Upon identifying the target, the guides return home to mobilize the worker robots to the target using chain formation.

\end{document}